\documentclass{article}
\PassOptionsToPackage{numbers,compress}{natbib}
\usepackage[final]{neurips_2025}

\usepackage[utf8]{inputenc} 
\usepackage[T1]{fontenc}    
\usepackage{hyperref}       
\usepackage{url}            
\usepackage{booktabs}       
\usepackage{amsfonts}       
\usepackage{nicefrac}       
\usepackage{microtype}      
\usepackage{xcolor}         

\usepackage{graphicx} 
\usepackage{natbib}
\usepackage{enumitem}
\usepackage{bbm}
\usepackage{mathtools}
\usepackage{algorithm}
\usepackage[noend]{algorithmic}

\usepackage{amsmath}
\usepackage{amsthm}
\usepackage{wasysym}
\usepackage{subcaption}
\usepackage{caption}
\usepackage{rotating}
\usepackage{adjustbox}
\usepackage{wrapfig}

\usepackage{booktabs} 
\usepackage{multirow}
\usepackage{array} 
\usepackage{makecell}

\newtheorem{definition}{Definition}

\newtheorem*{running_example}{Running Example}

\usepackage{listings}
\usepackage{url}
\PassOptionsToPackage{hyphens}{url}\usepackage{hyperref}
\usepackage{xspace}

\title{Ground-Compose-Reinforce: Grounding Language in Agentic Behaviours using Limited Data}

\author{
  Andrew Li, ~~Toryn Klassen$^\dagger$, ~~Andrew Wang, ~~Parand Alamdari$^\dagger$, ~~Sheila McIlraith$^\dagger$ \\
 Department of Computer Science, University of Toronto \\ Vector Institute for Artificial Intelligence\\ $^\dagger$ Schwartz Reisman Institute for Technology and Society \\
 Toronto, Canada \\
 \texttt{\{andrewli,toryn,andrewwang,parand,sheila\}@cs.toronto.edu} 
}

\date{February 2025}

\begin{document}

\newcommand{\showme}[1]{}

\newcommand{\notes}[1]{}

\newcommand{\headers}[1]{}

\newcommand{\consider}[1]{}   

\newcommand{\donotedit}[1]{\textcolor{red} {{\bf DO NOT EDIT THIS FILE. IT'S NOT THE VERSION WE'RE WORKING WITH}}}

\newcommand{\revisit}[1]{{\color{blue} #1}}
\newcommand{\added}[1]{{\color{red} #1}}
\newcommand{\alt}[1]{{\color{brown} $\backslash$#1}}
\newcommand{\altnoslash}[1]{{\color{brown} #1}}
\newcommand{\altsm}[1]{{}}

\newcommand{\removeDH}[1]{{\color{green} #1}}
\newcommand{\removeSM}[1]{{\color{green} #1}}
\newcommand{\removeTK}[1]{{\color{green} #1}}
\newcommand{\remove}[1]{}
\newcommand{\removemaybe}[1]{{\color{brown} #1}}
\newcommand{\removeifneeded}[1]{{\color{cyan} #1}}

\newcommand{\myhide}[1]{}

\newcommand{\todoanyone}[1]{\textcolor{orange}{({\bf TODO anyone:} #1)}}

 \newcommand{\nocommentsm}[1]{}
 \newcommand{\addCRC}[1]{}

 \newcommand{\dividelineit}[1]{\begin{center}-------------------------\end{center}}


\newif\ifcomments
\commentstrue

\ifcomments

    \newcommand{\commentsm}[1]{\textcolor{orange}{({\bf SM:} #1)\\}}
    \newcommand{\commentsmhide}[1]{}
    
    \newcommand{\commenttk}[1]{\textcolor{cyan}{({\bf TK:} #1)\\}}
    \newcommand{\commenttkhide}[1]{}
    
    \newcommand{\commental}[1]{\textcolor{magenta}{({\bf AL:} #1)}}
    \newcommand{\commentalhide}[1]{}

    \newcommand{\commentaw}[1]{\textcolor{red}{[AW: #1]}}
    \newcommand{\commentawhide}[1]{}

    \newcommand{\commenthc}[1]{\textcolor{green}{[HC: #1]}}
    \newcommand{\commenthchide}[1]{}

\else

    \newcommand{\commentsm}[1]{}
    \newcommand{\commentsmhide}[1]{}
    
    \newcommand{\commenttk}[1]{}
    \newcommand{\commenttkhide}[1]{}
    
    \newcommand{\commental}[1]{}
    \newcommand{\commentalhide}[1]{}

    \newcommand{\commentaw}[1]{}
    \newcommand{\commentawhide}[1]{}

    \newcommand{\commenthc}[1]{}
    \newcommand{\commenthchide}[1]{}
    
\fi


\maketitle

\begin{abstract}

    Grounding language in perception and action is a key challenge when building situated agents that can interact with humans, or other agents, via language. In the past, addressing this {challenge} has required manually designing the language grounding or curating massive datasets that associate language with the environment. We propose Ground-Compose-Reinforce, an end-to-end, neurosymbolic framework for training RL agents directly from high-level task specifications---
    {without manually designed reward functions or other domain-specific oracles, and without massive datasets}. These task specifications take the form of \emph{Reward Machines}, automata-based representations that capture high-level task structure and are in some cases autoformalizable from natural language. Critically, we show that Reward Machines can be grounded using limited data by exploiting {compositionality}. 
    Experiments in a custom Meta-World domain with only 350 labelled pretraining trajectories show that our framework faithfully elicits complex behaviours from high-level specifications---including behaviours that never appear in pretraining---while non-compositional approaches fail.

\end{abstract}
\section{Introduction}

Grounding language---connecting language with perception and action within an environment---is a fundamental challenge when building robots and other agents that are interfaced through language. One popular approach to addressing this challenge is to 
employ a \emph{manually-designed} domain-specific interpretation of language, 
such as a language-conditional reward function or success detector (e.g. \citep{chevalier2018babyai, hermann2017grounded, hill2020grounded, chaplot2018gated}). 
For instance, in the BabyAI benchmark \citep{chevalier2018babyai}, successful execution of instructions like ``go to the red ball'' can be evaluated programmatically in the environment simulator, providing a reward signal for learning language-conditioned behaviours. Such instances of grounded language generalize to arbitrary scenarios and controlled subsets of language by design, but are hard to hand-engineer in complex, non-simulated settings based on raw perceptual inputs like pixels.

The recent advent of large language models (LLMs) has inspired an alternative 
pathway to grounding language:
training on diverse datasets that pair language descriptions with environment trajectories (e.g., $\pi_0$ \citep{black2024pi_0}, RT-2 \citep{brohan2023rt}, LIV \citep{ma2023liv}, VPT \citep{baker2022video}). While this obviates the need for manually designed language groundings, 
it typically requires enormous datasets in order to capture the broad scope of language usage within an environment \citep{chevalier2018babyai, kaplan2020scaling, aghajanyan2023scaling}. For agentic applications that are data-intensive (e.g. robotics) or where access to trajectory data is limited, such data-driven language models are prone to failure on complex or out-of-distribution tasks \citep{StechlyNeurIPS2024thoughtlessness, shi2025hi, lifshitz2023steve, DBLP:journals/corr/abs-2108-07258}. 

\begin{figure}
\centering
\includegraphics[width=\textwidth]{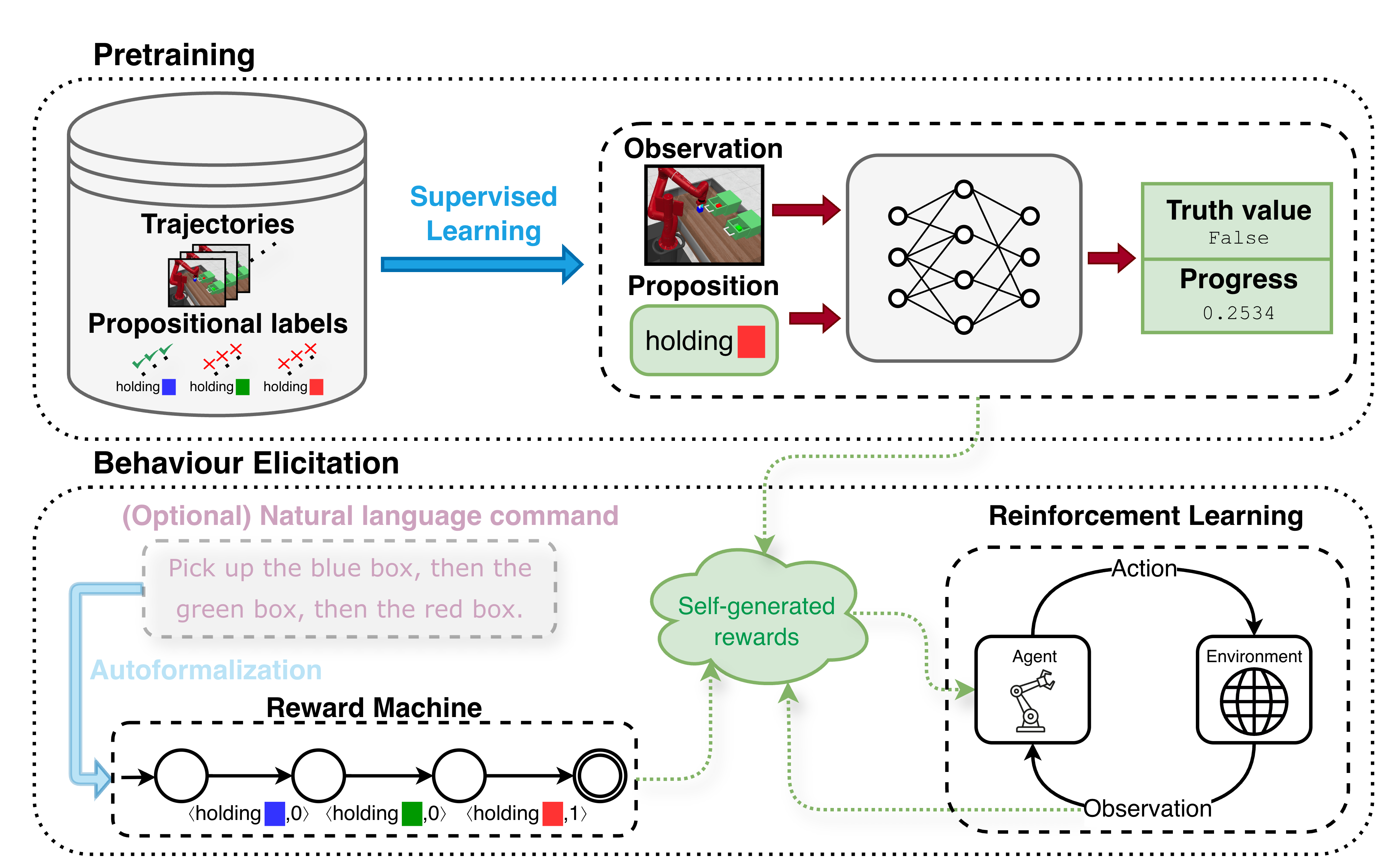}
\caption{Ground-Compose-Reinforce, a lightweight framework for training RL agents directly from Reward Machine specifications, without oracles like reward functions or feature detectors.
In pretraining, we learn to map propositional symbols to context-specific truth values (``is the robot holding the red block?'') and progress signals (``how close is the robot to holding the red block?''). To elicit behaviours, we prompt the agent via a Reward Machine composed of these symbols (or via natural language, if an autoformalizer is available). The agent then synthesizes its \emph{own} dense reward function and interacts with the environment to learn the desired behaviour via RL. }
\label{fig:gcr-framework}
\vspace{-1em}
\end{figure}

We propose \textbf{Ground-Compose-Reinforce}, an {end-to-end} framework for training reinforcement learning (RL) agents directly from high-level task specifications, \textit{without} relying on manually designed rewards or feature detectors. We represent tasks via \emph{Reward Machines} (RMs)~\cite{icarte2018using,icarte2022reward}, an automata-based task specification language
that exposes temporally extended task structure 
in terms of a set of atomic propositional symbols.  
Specifically, our framework: (1) \textbf{grounds} these atomic symbols in the environment by pretraining on a limited dataset of labelled trajectories; (2) \textbf{composes} these symbols via RMs to express complex tasks; and (3) for any such task, trains an agent to solve that task via \textbf{RL} on self-generated dense rewards based on its 
learned interpretation of the RM. 
Overall, Ground-Compose-Reinforce enables the elicitation of a wide range of temporally and logically structured behaviours expressed as RMs (specified directly or in some cases autoformalized from natural language), requires minimal pretraining data, and does not rely on external, domain-specific oracles 
for training or execution.
In this paper, we present the following novel contributions:
\begin{enumerate}[leftmargin=1em]
    \item A conceptual, end-to-end framework for {compositionally} grounding language in behaviours based on {Reward Machines} and reinforcement learning (Figure~\ref{fig:gcr-framework}). Critically, our framework requires minimal labelled trajectory data for pretraining, and does not require external oracles like reward functions, success detectors, or feature detectors.    
    \item A compositional reward shaping strategy for Reward Machine tasks that is critical for strong performance in Meta-World \citep{yu2020meta}. Our strategy addresses \emph{propositional sparsity}, where propositions of interest (e.g. ``pick up the block'') are rarely satisfied through random exploration.
    \item Experiments across diverse Reward Machine tasks, including temporally extended gridworld navigation and robotic manipulation in Meta-World. Our approach elicits diverse behaviours in Meta-World (including out-of-distribution behaviours that never appear in the pretraining dataset) from only 350 labelled pretraining trajectories while non-compositional approaches fail. 
    
\end{enumerate}

\section{Related Work}

\textbf{Grounded Language Learning.} Several past works have explored grounded language learning. \citet{hermann2017grounded} show that an RL agent in a 3-D environment can consistently navigate to target objects described via language by being rewarded for successful trajectories. \citet{hill2020grounded} show that an agent can learn new word-object bindings and apply that knowledge to solve tasks within the same episode. \citet{liu2023simple} show that a meta-RL agent indirectly learns to interpret language-based advice embedded within an environment to improve its task performance. \citet{chaplot2018gated} propose a neural architecture for grounded language learning in a 3-D Doom environment, considering both RL with success-based rewards and imitation learning from an oracle policy. \citet{chevalier2018babyai} propose the BabyAI platform for learning a synthetic language in a 2-D gridworld, finding that existing RL and imitation learning approaches are sample inefficient and generalize poorly. Unlike our approach, these methods require significant manual design of reward functions or oracle policies.

Recent works learn grounded language from data, without domain-specific manual design. \citet{black2024pi_0, brohan2023rt} and \citet{kim2024openvla} 
present vision-language-action models for robotics while \citet{baker2022video} and \citet{lifshitz2023steve} present models for playing MineCraft. \citet{bahdanau2018learning} and \citet{ma2023liv} learn language-conditioned reward functions from data. Works have also directly leveraged vision-language models for rewards in vision-based environments \citep{baumli2023vision, rocamonde2023vision, fu2024furl}, but typically do not leverage language compositionality and are prone to failure on complex or out-of-distribution tasks. \citet{shi2025hi, yuan2023plan4mc, huang2022inner} and \citet{ahn2022can} decompose complex tasks at execution time via language models. While this shares motivation with our work, we represent tasks via RMs and exploit compositional task structure for both training and execution.

\textbf{Formal Languages for RL.} Formal languages like Linear Temporal Logic (LTL) \citep{pnueli1977temporal} and other associated formal structures\footnote{Henceforth ``formal languages,'' for ease of exposition.} such as RMs \citep{icarte2022reward} have a rich history of application in the control \citep{karaman2008optimal, kloetzer2008fully}, verification \citep{baier2008principles, moon1992automatic, pnueli2005applications}, monitoring, and synthesis of dynamical systems. Recently, they have risen in popularity in deep RL for white-box specification of rich temporally extended reward functions \citep{icarte2018using, li2017reinforcement, camacho2019ltl, voloshin2023eventual} and in a number of cases can be automatically generated from natural language commands (e.g., \citep{brunello2019synthesis, liu2022lang2ltl, fuggitti2023nl2ltl, chen2023nl2tl}). Several works show that formal languages enable compositional generalization to unseen instructions. \citet{vaezipoor2021ltl2action, kuo2020encoding} and \citet{NEURIPS2024_858fc542} train instruction-conditioned policies that zero-shot generalize to unseen instructions by training on procedurally generated LTL formulas. \citet{NEURIPS2023_7b35a69f, liu2024skill, leon2021nutshell} and \citet{jackermeierdeepltl} train transferable skills that can be invoked by a planner.  \citet{tasse2024skill} consider how value functions and policies can be composed zero-shot for arbitrary RM tasks, but assume that all the tasks can be captured via a finite, predetermined set of goal states. Our compositional reward shaping approach also builds on prior methods. \citet{camacho2019ltl, furelos2021induction} and \citet{10.24963/kr.2024/85} provide potential-based rewards for RMs based on the current RM state, but such methods provide no signal when target propositions rarely occur. Several works propose continuous progress signals for each proposition \citep{li2017reinforcement, jiang2021temporal, elbarbari2021ltlf, jothimurugan2019composable, 8968254, 7799279}, but these approaches are limited to tasks with a binary success criterion. While progress signals are typically manually specified, we show how they can be learned directly from data.

Nearly all formal-language-based deep RL methods assume access to an external evaluator of symbolic features (a.k.a. a labelling function). To our knowledge, only a few works specifically avoid this assumption, but they instead depend on an external reward signal. \citet{li2024reward, li2022noisy} consider the implications of noisily grounding symbols when using RMs. \citet{HydeECAI2024triggers} and \citet{christoffersen2023learning} infer RMs from the reward signal as an inductive bias for RL. \citet{umili2024neural, umili2023grounding, kuo2020encoding, pmlr-v70-andreas17a} and \citet{pmlr-v70-oh17a} show that \textit{ungrounded} formal specifications can improve RL by providing information about the task structure. In contrast to these works, our approach does not rely on an external symbol evaluator or reward function.

\section{Preliminaries}

\subsection{Reinforcement Learning}
A reinforcement learning (RL) problem considers an environment modelled as a \emph{Markov Decision Process} (MDP) $\langle \mathcal{S}, \mathcal{A}, \mathcal{T}, \mathcal{P}, R, \mu, \gamma \rangle$, where $\mathcal{S}$ is a set of states, $\mathcal{A}$ a set of actions, $\mathcal{T} \subset \mathcal{S}$ a set of terminal states, $\mathcal{P}: \mathcal{S} \times \mathcal{A} \to \Delta \mathcal{S}$ a transition probability distribution, $R: \mathcal{S} \times \mathcal{A} \times \mathcal{S} \to \mathbbm{R}$ a reward function, $\mu \in \Delta \mathcal{S}$ an initial probability distribution, and $\gamma \in [0,1]$ a discount factor. An episode begins with $s_0 \sim \mu$, and at each time $t \ge 0$ the agent chooses an action $a_t$, then observes the next state $s_{t+1} \sim \mathcal{P}(s_t, a_t)$ and reward $r_{t+1} = R(s_t, a_t, s_{t+1})$, repeating until a terminal state $s_T \in \mathcal{T}$ is reached. We refer to full episodes as \emph{trajectories}, denoted by $\tau = \langle s_0, a_0, s_1, a_1, \ldots \rangle$. A \emph{history} $h_t = \langle s_0, a_0, s_1, a_1, \ldots s_t \rangle$ refers to the states and actions up to time $t$. 
The agent's goal is to interact with the environment to learn a policy  $\pi(a_t | s_t)$ that maximizes the \emph{expected discounted return} $\mathbb{E}_{\tau \sim \pi}[\sum_{t=1}^T {\gamma^tr_t}]$ (where the episode length $T$ can be $\infty$).


\begin{figure}
    \includegraphics[width=\textwidth]{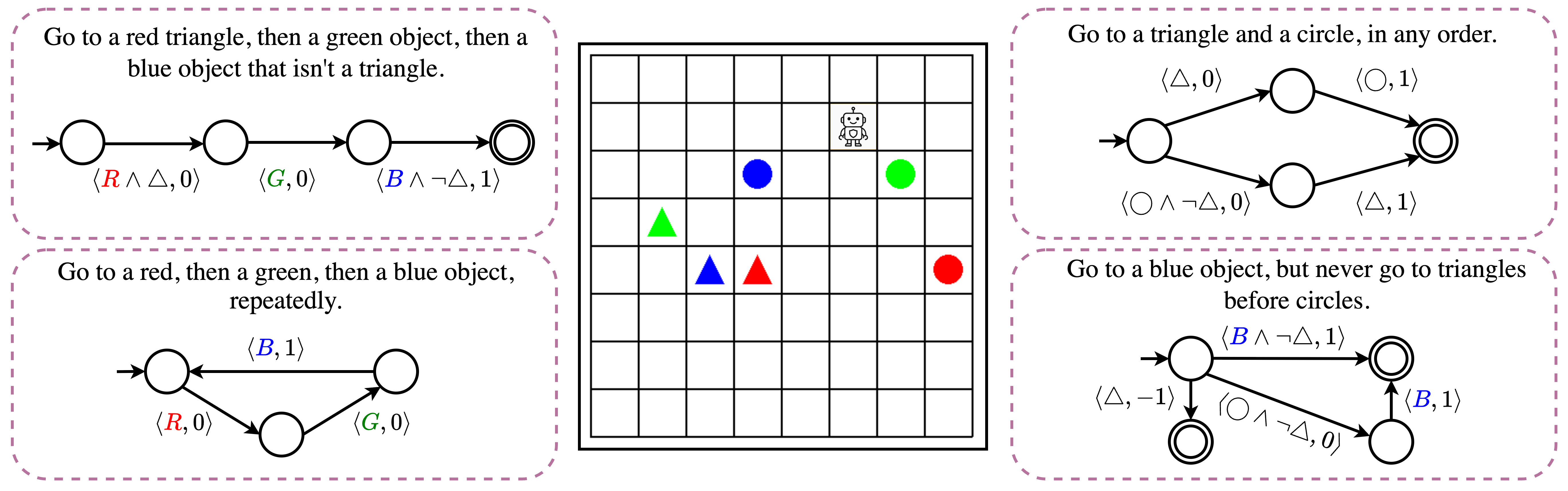}
    \caption{Four temporally extended tasks in a gridworld expressed as Reward Machines over the propositions $\mathcal{AP} = \{\mathrm{\textcolor{red}{R}}, \mathrm{\textcolor{green}G}, \mathrm{\textcolor{blue}B}, \triangle, \bigcirc \}$. An edge  labelled $\langle \varphi, r \rangle$ indicates the logical condition $\varphi$ for when the corresponding transition should be followed, and the reward $r$ that is yielded as a result. Doubled circles indicate terminal states, and we omit non-rewarding self-loop edges to aid readability.} 
    \vspace{-1em}
    \label{fig:rm_examples}
\end{figure}

\subsection{Task Specification via Reward Machines} 

Formal languages with sequential or temporal structure---including RMs \citep{icarte2022reward}, \textit{regular expressions} \citep{brafman2019regular}, and various \textit{temporal logics} \cite{7799279, vaezipoor2021ltl2action}---offer an intuitive and expressive interface for specifying tasks in RL while supporting representations that capture compositional task structure. Starting with a predefined set of \emph{atomic propositions} $\mathcal{AP}$ that represent abstract, binary features of environment states, such languages can be used to express temporally extended properties or reward functions.
In this work, we focus on tasks specified as RMs, which subsume several formal languages of interest \citep{camacho2019ltl} such as LTL over finite traces \citep{de2013linear, baier2006planning} and regular expressions. 

An RM is an automaton that captures the structure of a reward function over the abstract vocabulary $\mathcal{AP}$ (Figure~\ref{fig:rm_examples}). 
It takes as input a sequence of \emph{truth assignments} $\langle \omega_1, \omega_2, \ldots \rangle$ over $\mathcal{AP}$, where each $\omega_t \in 2^\mathcal{AP}$ denotes the subset of $\mathcal{AP}$ that holds true at time $t$, and outputs a corresponding sequence of rewards $\langle r_1, r_2, \ldots \rangle$. An RM has a finite set of internal states $\mathcal{U}$ and begins in a fixed state $u_0$ at time $t=0$. At each step $t \ge 1$, the RM updates its state to $u_t \in \mathcal{U}$ and emits a reward $r_t$ based on the current input $\omega_t$ and the previous state $u_{t-1}$. This continues until the RM enters a terminal state from a designated subset $\mathcal{F}\subset \mathcal{U}$.

\begin{running_example}
The gridworld in Figure~\ref{fig:rm_examples} will serve as a running example. Consider the top-left RM, which describes a task composed of three subgoals to be completed in a fixed order. When given an input sequence $\langle \omega_1, \omega_2, \ldots \rangle$ that identifies the values of \emph{all} propositions at each time $t$ (whether the agent is at a red object, a green object, a circle, and so on), the RM state $u_t \in \mathcal{U}$ tracks which subgoals have been completed and transitions to the next RM state as soon as the agent achieves the current subgoal. When a red triangle, a green object, and a blue object that isn't a triangle are reached in that order, the RM terminates with a reward of 1.
\end{running_example}

\begin{definition}
A \textbf{Reward Machine} $\mathcal{R}$ is defined as a tuple $\langle \mathcal{U}, u_0, \mathcal{F}, \mathcal{AP}, \delta_u, \delta_r \rangle$, where $\mathcal{U}$ is a finite set of states, with initial state $u_0 \in \mathcal{U}$ and terminal states $\mathcal{F} \subset \mathcal{U}$; $\mathcal{AP}$ is a set of propositions; $\delta_u: \mathcal{U} \times 2^\mathcal{AP} \to \mathcal{U}$ is a transition function that updates the RM state based on the current truth assignment; and $\delta_r: \mathcal{U} \times 2^\mathcal{AP} \to \mathbb{R}$ is a reward function that emits a reward at each step.
\end{definition}

As shown in Figure~\ref{fig:rm_examples}, the transition and reward functions of an RM can be intuitively and compactly specified via a set of labelled edges of the form $\langle u, u', \varphi, r \rangle$, indicating that if a truth assignment $\omega$ satisfies the formula $\varphi$ (denoted by $\omega \models \varphi$), then $\delta_u(u, \omega) = u'$ and $\delta_r(u, \omega) = r$.

\subsection{Grounded Interpretations} 
\label{sec:grounded_interpretations}

RMs express tasks in terms of abstract symbols (e.g., $\mathrm{\textcolor{red}{R}}$, $\triangle$), but these symbols must be \emph{grounded} in the environment to be meaningful. 
This is achieved via a \emph{labelling function} $\mathcal{L}: \mathcal{S} \to 2^\mathcal{AP}$ that maps each MDP state $s$ to the set $\omega \subseteq \mathcal{AP}$ of propositions that hold true in $s$. For an RM $\mathcal{R}$ over propositions $\mathcal{AP}$, any such $\mathcal{L}$ also grounds $\mathcal{R}$ in the environment by inducing a reward sequence for any MDP trajectory $\tau$: first, states $s_t$ in $\tau$ are converted into truth assignments $\omega_t = \mathcal{L}(s_t)$, then the RM is simulated over the sequence $\langle \omega_1, \omega_2, \ldots \rangle$ to generate a sequence of rewards until termination.



An \textit{RM-MDP} augments an environment (represented by a reward-free MDP) with a concrete reward function captured by an RM and labelling function. The resulting reward function is generally \emph{non-Markovian} with respect to MDP states $\mathcal{S}$, since the reward at time $t$ depends on the internal RM state $u_t$. While one might consider expressing optimal behaviours via a history-based policy $\pi(a_t | h_t)$, this is unnecessary if the agent has oracle access to $\mathcal{L}$ since RM-MDPs are Markovian over the extended state space $\mathcal{S} \times \mathcal{U}$ \citep{icarte2022reward}. The agent can use $\mathcal{L}$ to compute $\omega_t = \mathcal{L}(s_t)$ and recursively simulate the RM to track $u_t$. 
Thus, it is typical to express policies in the form $\pi(a_t | s_t, u_t)$, where the RM state $u_t$ compactly encodes the history $h_t$ and is sufficient for optimal decision making.


\begin{definition}
An \textbf{RM-MDP} is a triple $\langle \mathcal{M}, \mathcal{R}, \mathcal{L} \rangle$, where $\mathcal{M} = \langle \mathcal{S}, \mathcal{A}, \mathcal{P}, \mu, \gamma \rangle$ is an MDP without rewards or terminal states, $\mathcal{R} = \langle \mathcal{U}, u_0, \mathcal{F}, \mathcal{AP}, \delta_u, \delta_r \rangle$ is an RM, and $\mathcal{L}: \mathcal{S} \to 2^\mathcal{AP}$ is a labelling function. The RM-MDP is equivalent to an MDP with state space $\mathcal{S} \times \mathcal{U}$ and reward function induced by $\mathcal{R}$ and $\mathcal{L}$.
\end{definition}

\begin{running_example}
The RM at the top left of Figure~\ref{fig:rm_examples} captures the high-level structure of a multi-stage task. To map environment trajectories into concrete rewards for this RM, we need a \emph{labelling function} $\mathcal{L}: \mathcal{S} \to 2^\mathcal{AP}$ that connects abstract propositions like $\mathrm{\textcolor{red}{R}}$ and $\triangle$ to environment states.
\end{running_example}

\section{Problem Setting}

Our goal in this work is to faithfully elicit behaviours from an agent given only a high-level task specification in the form of an RM, $\mathcal{R}$ such as the ones depicted in Figure~\ref{fig:rm_examples}. 
$\mathcal{R}$ can be specified directly, translated from other formal languages like LTL \citep{camacho2019ltl}, generated from a symbolic planner \citep{illanesYTM2019symbolic,illanesYTM2020symbolic}, or sometimes autoformalized from natural language. Formally, we consider an environment $\mathcal{M} = \langle \mathcal{S}, \mathcal{A}, \mathcal{P}, \mu, \gamma \rangle$ (an MDP without rewards or terminal states) and a finite set of propositional symbols $\mathcal{AP}$. For any given RM task $\mathcal{R} = \langle \mathcal{U}, u_0, \mathcal{F}, \mathcal{AP}, \delta_u, \delta_r \rangle$, we wish to obtain a policy $\pi_\mathcal{R}(a_t | h_t)$ that performs well in the RM-MDP $\langle \mathcal{M}, \mathcal{R}, \mathcal{L}^* \rangle$, where $\mathcal{L}^*: \mathcal{S} \to 2^\mathcal{AP}$ reflects a ground-truth interpretation of the propositions $\mathcal{AP}$ in the environment.\footnote{For the problem statement, we consider the more general form of history-based policies, rather than policies conditioned on the ground-truth RM state, which depend on $\mathcal{L}^*$.}

\textbf{Assumptions.} We aim to obtain $\pi_\mathcal{R}$ \emph{without} online access to $\mathcal{L^*}$ or to an external reward function that evaluates ground-truth performance with respect to $\mathcal{R}$.\footnote{Such oracles can be notoriously hard to design in practice \citep{clark2016faulty} and often require internal simulator access.}  
In order to connect symbols $\mathcal{AP}$ with environment percepts, we instead assume access to a fixed pretraining dataset $\mathcal{D} = \{\langle \tau^i, \omega^i \rangle\}_{i=1}^N$ of trajectories $\tau^i = \langle s^i_0, a^i_0, s^i_1, \ldots \rangle$ with corresponding labels $\omega^i = \langle \mathcal{L}^*(s^i_0), \mathcal{L}^*(s^i_1), \ldots \rangle$. In practice, such labels can be obtained via crowdsourced annotations \cite{stiennon2020learning} or self-supervised learning \cite{radford2021learning}. 

For a task $\mathcal{R}$, the agent is allowed an arbitrary number of interaction episodes with the environment before committing to a final policy $\pi_\mathcal{R}$. However, during this interaction phase, the agent must learn in a self-supervised manner as the environment does not provide a separate reward signal. 

\section{Ground-Compose-Reinforce}
\label{sec:gcr}

We propose an end-to-end framework for this setting called \emph{Ground-Compose-Reinforce} (Figure~\ref{fig:gcr-framework}). In the pretraining phase, the agent first {grounds} propositional symbols $\mathcal{AP}$ in environment states via supervised learning on $\mathcal{D}$. In the behaviour elicitation phase, the agent is given a task as an RM $\mathcal{R}$ composed over symbols $\mathcal{AP}$. The agent then learns a policy $\pi_{\mathcal{R}}$ by interacting with the environment and synthesizing its own learning signal for RL based on $\mathcal{R}$ and its learned interpretation of $\mathcal{AP}$. We hypothesize that this bottom-up approach to grounding language in behaviours---first learning the meanings of individual symbols and then composing them to interpret complex tasks---is an effective strategy. The remainder of this section describes the core implementation of Ground-Compose-Reinforce and in Section~\ref{sec:sparse_rewards}, we raise and address an issue called \emph{propositional sparsity} where the agent fails to learn in extremely long-horizon tasks.

\renewcommand{\algorithmicrequire}{\textbf{Input:}}
\renewcommand{\algorithmicensure}{\textbf{Output:}}

\begin{wrapfigure}[16]{r}{0.60\textwidth}
\vspace{-2.7ex}
\begin{minipage}{0.6\textwidth}
\begin{algorithm}[H]  
\caption{Ground-Compose-Reinforce for RMs}
\label{alg:reinforce}
\begin{algorithmic}[1]
\small
\REQUIRE MDP $\mathcal{M}$ without rewards, Propositional symbols $\mathcal{AP}$, Dataset $\mathcal{D}$ of labelled trajectories, RM task $\mathcal{R}$ over $\mathcal{AP}$

\COMMENT{Pretraining phase}
\STATE Train labelling function $\hat{\mathcal{L}}(s)$ on $\mathcal{D}$ using any binary classification method

\COMMENT{Behaviour elicitation phase}
\STATE Initialize policy $\pi_{\mathcal{R}}(a \mid s, u)$ arbitrarily
\FOR{each episode}
    \STATE Observe initial state $s$ in $\mathcal{M}$; set $u$ to the initial state of $\mathcal{R}$
    \WHILE{$u$ is non-terminal}
        \STATE Sample action $a \sim \pi_{\mathcal{R}}(\cdot \mid s, u)$
        \STATE Execute $a$ in $\mathcal{M}$ and observe next state $s'$
        \STATE Compute truth assignment $\hat{\omega} \gets \hat{\mathcal{L}}(s')$
        \STATE Update RM: $u' \gets \delta_u(u, \hat{\omega})$, $r \gets \delta_r(u, \hat{\omega})$
        \STATE Update $\pi_{\mathcal{R}}$ with RL for transition $\langle s, u, a, r, s', u' \rangle$
        \STATE Set $s \leftarrow s'$, $u \leftarrow u'$
    \ENDWHILE
\ENDFOR
\end{algorithmic}
\end{algorithm}
\end{minipage}
\end{wrapfigure}

\textbf{Core Algorithm.}
We ground the propositional symbols $\mathcal{AP}$ by learning a labelling function $\hat{\mathcal{L}}(s) \approx \mathcal{L}^*(s)$ via any binary classification method on $\mathcal{D}$. Given an RM task $\mathcal{R}$ over $\mathcal{AP}$, we solve a \emph{surrogate} RM-MDP $\langle \mathcal{M}, \mathcal{R}, \hat{\mathcal{L}} \rangle$ via RL. This surrogate task approximates the true RM-MDP $\langle \mathcal{M}, \mathcal{R}, \mathcal{L}^* \rangle$ for which the agent lacks supervision. Since the agent can query $\hat{\mathcal{L}}$ freely, it can simulate rewards $\langle {r}_1, {r}_2, \ldots \rangle$ and RM states $\langle {u}_1, {u}_2, \ldots \rangle$ for any trajectory, as described in Section~\ref{sec:grounded_interpretations}. Finally, we use these self-generated signals to train a policy $\pi_\mathcal{R}(a_t | s_t, u_t)$ as outlined in Algorithm~\ref{alg:reinforce}.



\begin{running_example}
Suppose our gridworld agent is expected to solve arbitrary RM tasks over the vocabulary $\{\mathrm{\textcolor{red}{R}}, \mathrm{\textcolor{green}G}, \mathrm{\textcolor{blue}B}, \triangle, \bigcirc \}$.
With Ground-Compose-Reinforce, the agent first connects these symbols to environment states via $\mathcal{D}$ (i.e. it learns to identify which states have red shapes, which states have triangles, etc). A human can then specify a new task as an RM composed over these symbols (e.g. ``go to a triangle and a circle, in any order'') without needing to program a task-specific reward function. The agent systematically evaluates its own performance on this task by composing its learned interpretations of $\{\mathrm{\textcolor{red}{R}}, \mathrm{\textcolor{green}G}, \mathrm{\textcolor{blue}B}, \triangle, \bigcirc \}$, providing a learning signal for RL.

\end{running_example}

\section{Compositional Reward Shaping} 
\label{sec:sparse_rewards}

\begin{wrapfigure}{r}{0.5\textwidth}
    \centering
    \vspace{-2.4ex}
    \includegraphics[width=0.5\textwidth]{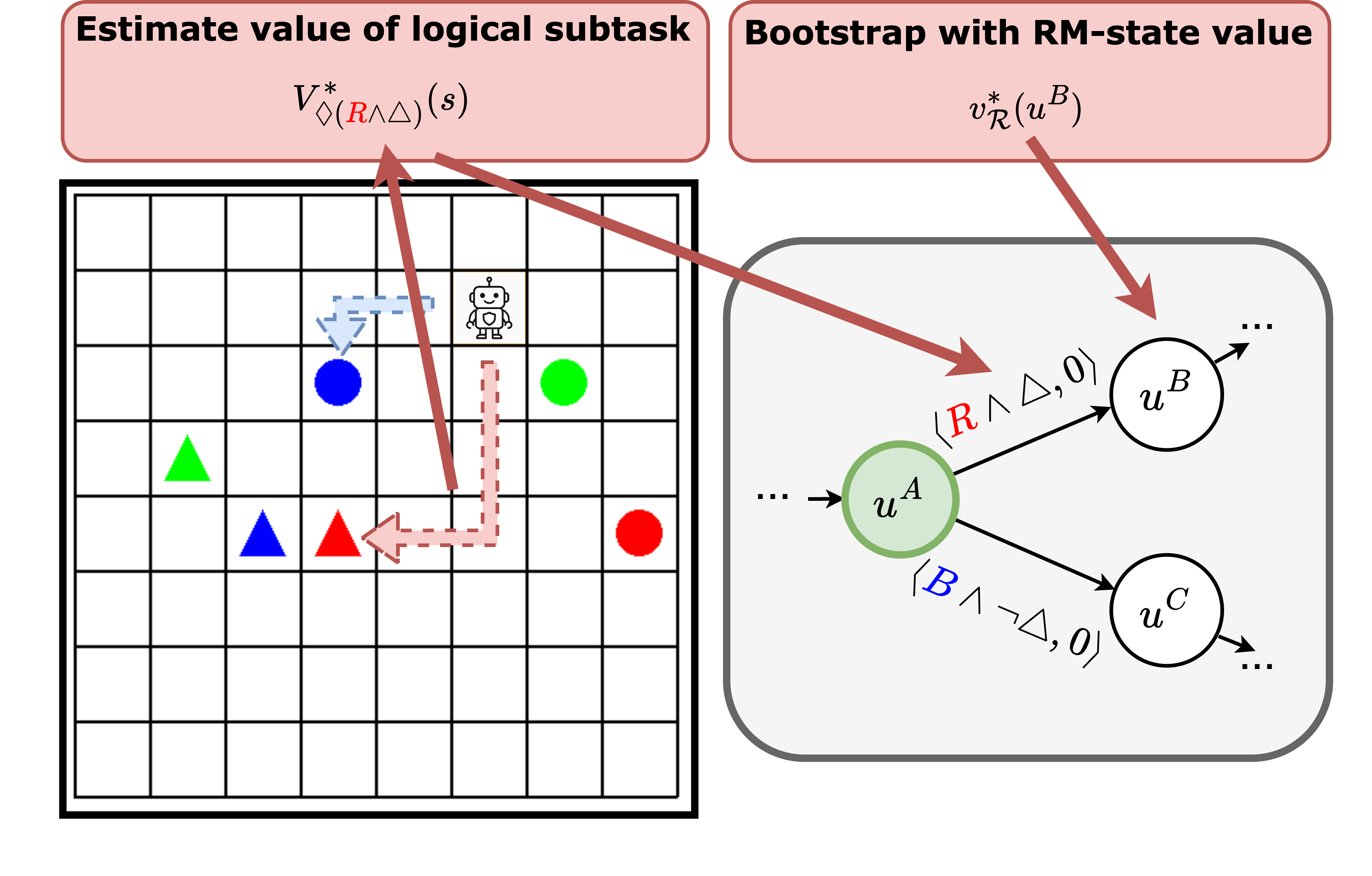}
    \vspace{-2em}
    \caption{An illustration of how we estimate optimal values in an RM-MDP. Suppose the agent is currently in RM state $u^A$ (green and bolded). To evaluate the expected return for the transition $u^A \to u^B$, we estimate how close the agent is to satisfying the formula on the transition (reaching the red triangle), and bootstrap with a coarse value estimate for RM state $u^B$. The overall value of $u^A$ is approximated by the maximum expected return across all outgoing transitions from $u^A$.  
     }
     \vspace{-1em}
     \label{fig:reward-shaping}
\end{wrapfigure}

An issue that can arise with the core implementation is \emph{sparse rewards}. While prior RM works have proposed additional learning signals for transitioning in the RM \citep{camacho2019ltl, furelos2021induction, 10.24963/kr.2024/85}, these methods fail when RM transitions \emph{themselves} pose an exploration challenge. This can be caused by \emph{propositional sparsity}: when $\mathcal{L}^*(s)$ is constant across most states under random exploration, the agent struggles to learn to meaningfully affect propositional values (e.g. consider a robot manipulation task that involves satisfying the proposition ``the box is picked up'').

To address this, we extend Ground-Compose-Reinforce with \emph{potential-based reward shaping} \citep{ng1999policy}. Specifically, for a given RM task $\mathcal{R}$, we estimate the \emph{optimal value function} (OVF) for the surrogate RM-MDP $\langle \mathcal{M}, \mathcal{R}, \hat{\mathcal{L}} \rangle$ by decomposing $\mathcal{R}$ into simpler subtasks. The most basic subtasks correspond to satisfying individual propositions or their negations, and we estimate the OVFs for these $2|\mathcal{AP}|$ tasks during pretraining via offline RL on $\mathcal{D}$. These $2|\mathcal{AP}|$ OVFs then serve as building blocks that can be composed to approximate the OVF for \emph{any} of the \emph{infinitely many} RM tasks over $\mathcal{AP}$.

\subsection{Deconstructing RMs into Logical Subtasks}
\label{subsec:logical_subtasks}

To estimate the OVF of the surrogate RM-MDP $\langle \mathcal{M}, \mathcal{R}, \hat{\mathcal{L}} \rangle$, denoted $V^*_\mathcal{R}(s,u)$, we combine two key ideas:
(1) treating each RM transition in $\mathcal{R}$ as an independent subtask, and (2) bootstrapping from a state-independent value function $v^*_\mathcal{R}(u)$. Consider the example in Figure~\ref{fig:reward-shaping}, where the agent is in MDP state $s$ and RM state $u^A$. Each RM transition is associated with a logical condition $\varphi$ (e.g., $\textcolor{red}{R} \land \triangle$ or $\textcolor{blue}{B} \land \lnot \triangle$), and we treat the satisfaction of $\varphi$ as its own subtask. Definition~\ref{def:pl} formalizes this: we introduce the class $\Diamond\mathbf{PL}(\mathcal{AP})$ of reachability tasks over propositional formulas, where each task $\Diamond \varphi$ entails reaching a state $s_T$ such that $\hat{\mathcal{L}}(s_T) \models \varphi$, terminating with reward 1 upon satisfaction. $\Diamond \varphi$ is itself an RM-MDP with a single non-terminal RM state (and a terminal RM state), and is therefore Markovian over $\mathcal{S}$. 
We denote its OVF as $V^*_{\Diamond \varphi}(s)$.

\begin{definition}
\label{def:pl}
For any propositional logic formula $\varphi$ over $\mathcal{AP}$, define $\Diamond \varphi$ as the task of reaching a state $s_T \in \mathcal{S}$ such that $\hat{\mathcal{L}}(s_T) \models \varphi$. The episode terminates and yields a reward of 1 if such a state is reached, and continues indefinitely with 0 reward otherwise. Let $\Diamond\mathbf{PL}(\mathcal{AP}) =\allowbreak \left\{ \Diamond \varphi \;\middle|\; \varphi \text{ is a propositional formula over } \mathcal{AP} \right\}$ denote the set of such reachability tasks.
\end{definition}

The second component of our method is to bootstrap using a state-independent value function $v^*_\mathcal{R}(u)$ that approximates the expected return from any RM state $u$ while ignoring the MDP state. We estimate $v^*_\mathcal{R}(u)$ using a variant of Value Iteration over the RM graph, following \citet{camacho2019ltl}. 
Finally, to estimate $V^*_\mathcal{R}(s,u)$, we evaluate each outgoing transition $\langle u, u', \varphi, r \rangle$ from $u$ by combining the subtask value $V^*_{\Diamond \varphi}(s)$ with the bootstrapped value of the next RM state $v^*_\mathcal{R}(u')$. The final estimate is the maximum expected return over all such transitions:
\begin{equation}
    V^*_\mathcal{R}(s,u) \approx \max_{\langle u, u', \varphi, r \rangle} \bigl[ V^*_{\Diamond \varphi}(s) \cdot \bigl(r + \gamma v^*_\mathcal{R}(u')\bigr) \bigr]
    \label{eq:value_rm}
\end{equation}
This approximation assumes no RM self-transitions with non-zero rewards. Appendix~\ref{sec:composition_supp} further justifies and explains our approximation while extending it to arbitrary RMs. 


\subsection{Further Deconstructing Logical Subtasks}

Approximation~\ref{eq:value_rm} allows us to estimate $V^*_\mathcal{R}(s,u)$ for any RM task $\mathcal{R}$ over $\mathcal{AP}$, provided we can estimate $V^*_{\Diamond \varphi}(s)$ for any $\varphi$ in $\Diamond \mathbf{PL}(\mathcal{AP})$. However, the number of propositional formulas over $\mathcal{AP}$ (up to logical equivalence) is $2^{2^{|\mathcal{AP}|}}$ and modelling a separate OVF for each such task is intractable. To address this, we further decompose logical formulas based on their structure. Any formula $\varphi$ can be rewritten in disjunctive normal form, i.e., as a disjunction of conjunctions of literals (where a literal is either a proposition $x$ or its negation $\lnot x$). We then approximate the OVF of $\Diamond \varphi$ using the semantics of fuzzy logic \citep{goguen1969logic}, where $\max$ represents disjunction and $\min$ represents conjunction.\footnote{Fuzzy operators have previously been applied for satisfaction of a temporal formula over quantitative signals \citep{li2017reinforcement, 8968254, 7799279}. While such tasks are binary in nature, we consider RMs, which can express other reward structures.} 


Let $\varphi = \xi_1 \lor \ldots \lor \xi_k$, where each $\xi_i$ is a conjunction of literals. We approximate:
\begin{equation}
    V^*_{\Diamond \varphi}(s) \approx \max_{i=1,\ldots,k} V^*_{\Diamond \xi_i}(s)
    \label{eq:disjunction}
\end{equation}
\noindent For each conjunctive clause $\xi = l_1 \land \ldots \land l_k$, where each $l_i$ is a literal, we approximate:
\begin{equation}
    V^*_{\Diamond \xi}(s) \approx \min_{i=1,\ldots,k} V^*_{\Diamond l_i}(s)
    \label{eq:conjunction}
\end{equation}
By composing Approximations~\ref{eq:value_rm}–\ref{eq:conjunction}, we can estimate the OVF for any RM task based on only $2|\mathcal{AP}|$ OVFs---namely, those for $\Diamond x$ and $\Diamond \lnot x$, for each $x \in \mathcal{AP}$. We refer to these as the \emph{primitive value functions} (PVFs).



\subsection{Final Remarks}

In this section, we showed that the optimal value function (OVF) of \emph{any} RM task $\mathcal{R}$ can be approximated using just $2|\mathcal{AP}|$ \emph{primitive value functions} (PVFs). From these $2|\mathcal{AP}|$ PVFs, we can estimate OVFs for \emph{doubly exponentially many} logical tasks ($2^{2^n}$) and \emph{infinitely many} RM tasks.
Each PVF quantifies progress toward satisfying a single proposition or its negation, and can be learned directly from $\mathcal{D}$ using any offline RL algorithm. 
One might view this approach as trading off expressivity for modularity: directly modelling the OVFs of all RM tasks is infeasible, so we instead model a small, reusable set of $2|\mathcal{AP}|$ PVFs at the cost of introducing some approximation error.  

Leveraging that we can estimate the OVF for any RM task $\mathcal{R}$, we extend Algorithm~\ref{alg:reinforce} with potential-based reward shaping to address propositional sparsity. Further details on learning PVFs, sources of approximation error, and the potential-based reward shaping scheme are provided in Appendix~\ref{sec:composition_supp}. 

\section{Experiments}
\label{sec:experiments}

We conducted experiments to evaluate the following research questions:
\begin{enumerate}[label=\textbf{RQ\arabic*},itemsep=0.2em, topsep=0em, parsep=0pt, leftmargin=3em]
    \item \textbf{Grounding RMs in Behaviours}: With Ground-Compose-Reinforce, can we faithfully elicit behaviours given high-level task specifications (RMs)?

    \item \textbf{Compositional Generalization}: Can we elicit meaningful, out-of-distribution (OOD) behaviours beyond those observed in $\mathcal{D}$?
    
    \item \textbf{Propositional Sparsity}: Can the agent operate in extremely long-horizon environments where propositional values are hard to alter with random exploration? 
\end{enumerate}

\emph{Code/videos available at:}  \url{https://github.com/andrewli77/ground-compose-reinforce}.

\subsection{Experimental Setup}

\begin{wrapfigure}{r}{0.38\textwidth}
    \centering
    \vspace{-2.8ex}
    \includegraphics[width=0.38\textwidth]{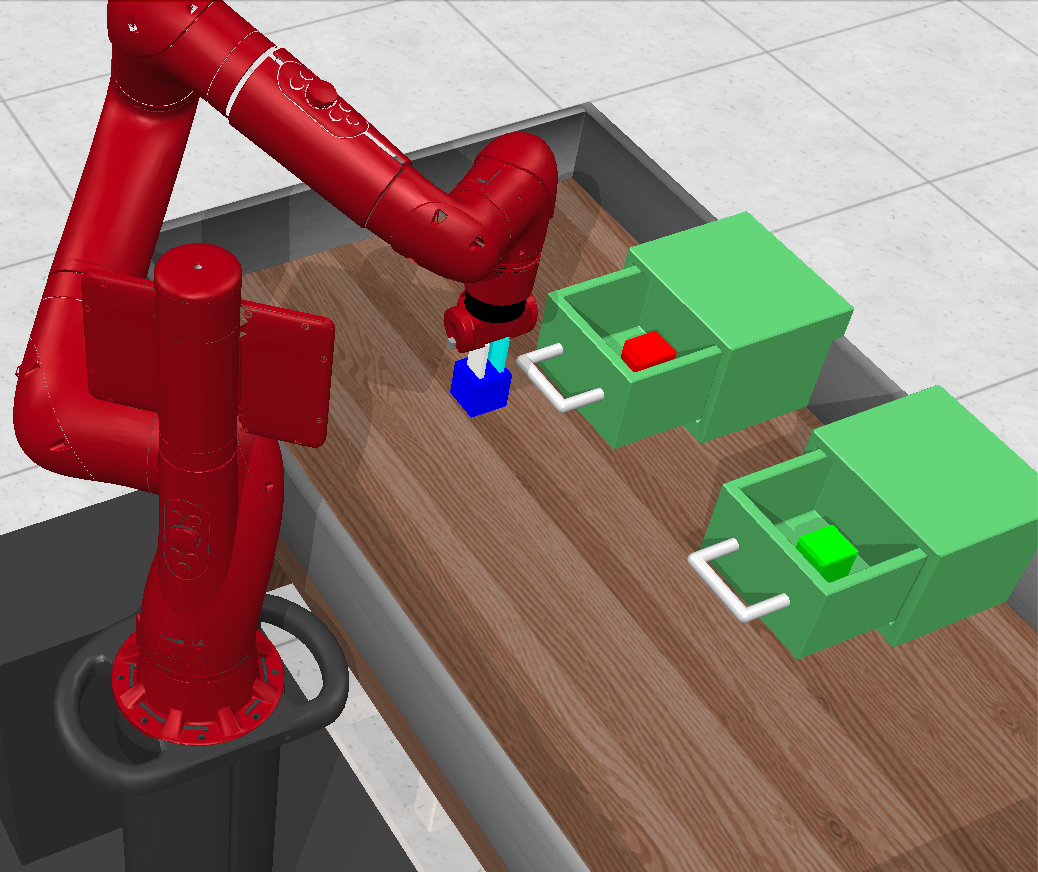}
    \caption{\textit{DrawerWorld} is a custom Meta-World environment where the agent can interact with two drawers and three boxes. Propositions capture whether: each drawer is open; each box is lifted by the agent; a given box is in a given drawer.}
    \vspace{-3.5ex}
    \label{fig:envs}
\end{wrapfigure}

We considered two domains: an image-based gridworld with randomized object locations called \emph{GeoGrid} (introduced in the running example), and a Meta-World-based robotics environment called \emph{DrawerWorld} (Figure~\ref{fig:envs}). Full details on the setup can be found in Appendix~\ref{sec:domain_details}.

\textbf{Pretraining Datasets.} We collected $\mathcal{D}$ under minimal assumptions about downstream tasks. In GeoGrid, $\mathcal{D}$ contains 5000 trajectories generated by a random policy. In DrawerWorld, we manually operated the robot to collect 350 trajectories involving generic behaviours (e.g., opening drawers, lifting boxes). To evaluate OOD generalization, we constrained $\mathcal{D}$ to only contain trajectories interacting with at most one box in DrawerWorld. Finally, trajectories were labelled using a handcrafted labelling function.

\textbf{Tasks.} We designed a diverse set of RM tasks (Table~\ref{tab:tasks}) that target behaviours rarely or never seen in $\mathcal{D}$. The GeoGrid tasks evaluate whether the agent can produce fine-tuned behaviours beyond the random-action trajectories observed in $\mathcal{D}$. The DrawerWorld tasks evaluate whether the agent can solve complex manipulation tasks that require composing behaviours observed in $\mathcal{D}$ (e.g., \emph{Pickup-Each-Box} requires handling all three boxes, while trajectories in $\mathcal{D}$ interact with at most one box). 

\subsection{Method and Baselines}

We benchmarked Ground-Compose-Reinforce (GCR) against several non-compositional baselines. Methods based on online RL (GCR, Bespoke Reward Model) use PPO \citep{schulman2017proximal} to train a policy from scratch. During execution, GCR captures memory via the RM state while all non-RM-based baselines encode the observation history using an additional GRU \citep{cho2014learning}. See Appendix~\ref{sec:implementation_details} for full implementation details and Appendix~\ref{sec:training_details} for full training details.

\textbf{Ground-Compose-Reinforce (ours).}  
We implemented GCR with potential-based reward shaping as described in Sections~\ref{sec:gcr} {and}~\ref{sec:sparse_rewards}. Both the predicted labelling function $\hat{\mathcal{L}}$ and PVFs are neural networks trained on $\mathcal{D}$ via supervised learning and offline RL, respectively.


\textbf{LTL-conditioned Behaviour Cloning (LTL-BC)} is a neural network policy $\pi(a_t | h_t, \varphi)$ that directly maps LTL specifications $\varphi$ to behaviours. We labelled each trajectory $\tau$ in $\mathcal{D}$ with an LTL description $\varphi$ (based on the propositional labels for $\tau$), then trained the policy to maximize the log-likelihood of actions in $\tau$, conditioned on the history $h_t$ and $\varphi$. For each downstream task in Table~\ref{tab:tasks}, we prompted the policy with an LTL formula that aligns with that task. 

We also trained bespoke models with advance knowledge of the downstream RM tasks. \textbf{Bespoke Reward Model} directly predicts rewards, optimal values, and terminations for all downstream tasks simultaneously. We labelled each trajectory in $\mathcal{D}$ with ground-truth rewards and terminations for each task based on the propositional labels, then trained the model to directly predict these quantities given the history $h_t$. Value estimates were trained via offline RL in a similar manner to GCR. Finally, we trained a policy via RL while using the learned value function for potential-based reward shaping. \textbf{Bespoke Behaviour Cloning (BC)} is a neural network policy that directly imitates successful trajectories in $\mathcal{D}$ for every downstream task. Due to the limited number of reward-worthy trajectories in $\mathcal{D}$, we considered any trajectory achieving positive return on that task as successful.

\begin{table}[t]
  \centering
  \small
  \caption{List of RM tasks. For each, we report the mean ($\mu_\mathcal{D}$) and max ($\max_\mathcal{D}$) undiscounted return over trajectories in $\mathcal{D}$, along with the max achievable expected return of any policy (Max; if unknown, we report the highest average return observed in our experiments). {Some tasks involve behaviours that are rarely or never observed in $\mathcal{D}$}.
  }
  \label{tab:tasks}
  \renewcommand{\arraystretch}{1.0}
  \begin{tabular}{p{0.17\textwidth}p{0.52\textwidth}ccc}
    \toprule
    \textbf{Task} & \textbf{Description} & 
    \multicolumn{3}{c}{\textbf{Return}} \\ 
    \midrule
    \textit{GeoGrid} & & $\mu_\mathcal{D}$ & $\mathrm{max}_\mathcal{D}$ & Max \\
    \midrule
    Sequence & Go to a red $\triangle$, then a green $\triangle$, then a blue $\triangle$. & $0.04$ & $1$ & $1$ \\
    Loop & Repeatedly go to a red $\triangle$, then a green $\triangle$, then a blue $\triangle$. & $0.04$ & $3$ & $5.36$ \\
    Logic & Go to all six objects, but always go to red objects before blue objects, and blue
objects before green objects. & $0.00$ & $1$ & $1$ \\
    Safety & Go to a red object, then a blue object, then a green object, but always
avoid $\triangle$. & $\!-\!0.84$ & $1$ & $1$ \\
    \midrule
    \multicolumn{3}{l}{\textit{DrawerWorld}} \\
    \midrule
    Hold-Red-Box & Lift and hold the red box as long as possible. & $41.7$ & $736$ & $1538$ \\
    Pickup-Each-Box & Pick up the red box, then the blue box, then the green box. & $0$ & $0$ & $1$ \\
    Show-Green-Box & Reveal the green box if it's in a closed drawer, then lift it. & $0.22$ & $1$ & $1$ \\
    \bottomrule
  \end{tabular}
  \vspace{-2em}
\end{table}



We also compared various reward shaping schemes for GCR. $\textbf{No RS}$ directly uses RM rewards without any reward shaping (i.e. Algorithm~\ref{alg:reinforce}). \textbf{High-Level RS}, inspired by \citet{camacho2019ltl}, uses a potential function that only considers the current RM state, but not the current MDP state.

\subsection{Results}

We ran each method's training pipeline five times and report the average final performances in Table~\ref{tab:results_rl}.
Performance was measured by {undiscounted return} (averaged over 100 evaluation episodes for GeoGrid or 20 for DrawerWorld), where rewards are with respect to the {ground-truth} labelling function $\mathcal{L}^*$. In Appendix~\ref{sec:learning_curves}, we report RL learning curves, both with respect to the agent's own reward model (without shaping rewards) and ground-truth rewards under $\mathcal{L}^*$. 

\emph{GCR consistently solves novel tasks, even when no successful demonstrations exist in $\mathcal{D}$.}  
On Loop, Hold-Red-Box and Pickup-Each-Box, it significantly outperforms even the best trajectories in $\mathcal{D}$. We attribute this success to GCR's ability to learn transferable knowledge from $\mathcal{D}$ and apply it towards novel task compositions, while fine-tuning behaviours with (self-supervised) RL. 


\emph{All non-compositional baselines fail to reliably solve any task.} Results show that LTL-BC, Bespoke Reward Model, and Bespoke BC do not fare well with limited pretraining trajectories. Learning curves show that the Bespoke Reward Model assigns near-zero rewards to most trajectories in GeoGrid, likely due to the rarity of positive demonstrations in $\mathcal{D}$. In DrawerWorld, it produces misaligned rewards, leading to {reward hacking} (evidenced by high returns under the learned reward model but low returns under the ground-truth $\mathcal{L}^*$).



\emph{Reward shaping enables long-horizon RL.}  
Our reward shaping strategy yields modest improvements in GeoGrid, but is critical to success in DrawerWorld, where behaviours like opening a drawer and picking up a box are nearly impossible to discover from random exploration alone. 


We conclude the following. GCR faithfully elicits behaviours from RM specifications, outperforming non-compositional approaches (\textbf{RQ1}). Moreover, GCR compositionally generalizes to OOD behaviours beyond those observed in $\mathcal{D}$ (\textbf{RQ2}). Finally, our compositional reward shaping strategy for GCR enables RL in long-horizon settings involving propositional sparsity (\textbf{RQ3}).

\begin{table}[t]
\small
\centering
\caption{Comparison of methods for eliciting behaviours from high-level task specifications. We report performance (undiscounted return with respect to ground-truth rewards) averaged over 5 runs with standard error.}
\label{tab:results_rl}
\renewcommand{\arraystretch}{1.0}
\begin{tabular}{@{}l@{\hskip 5pt}c@{\hskip 5pt}c@{\hskip 5pt}c@{\hskip 5pt}c@{\hskip 5pt}c@{\hskip 5pt}c@{}}
\toprule
\multirow{2}{*}{\textbf{Task}} 
& \multirow{2}{*}{\centering\textbf{GCR (Ours)}} 
& \multirow{2}{*}{\centering{LTL-BC}} 
& \multicolumn{1}{c}{{Bespoke}} 
& \multicolumn{1}{c}{{Bespoke}} 
& \multicolumn{1}{c}{\centering{GCR (Ours)}} 
& \multicolumn{1}{c}{\centering{GCR (Ours)}} \\
& & & {Reward Model} & {BC} & No RS & High-Level RS\\
\midrule
\multicolumn{6}{l}{\textit{GeoGrid}} \\
\midrule
Sequence         & $\textbf{1.00} \pm 0.00$ & $0.04 \pm 0.01$ & $0\pm0$            & $0.05\pm0.01$        & $0.94 \pm 0.03$ & $\textbf{1.00} \pm 0.00$ \\
Loop             & $\textbf{5.36} \pm 0.08$ & $0.03 \pm 0.01$ & $0\pm0$            & $0.04\pm0.01$        & $4.68\pm0.05$   & $5.27 \pm 0.08$ \\
Logic            & $\textbf{0.94}\pm0.01$ & $0 \pm 0$  & $0\pm 0$           & $0\pm0$              & $0.00 \pm 0.00$ & $\textbf{0.94} \pm 0.01$ \\
Safety           & $\textbf{1.00}\pm0.00$ & $\!-\!0.84 \pm 0.01$  & $\!-\!0.14 \pm 0.01$   & $\!-\!0.85\pm0.01$       & $0.23\pm0.11$   & $0.97\pm0.01$ \\
\midrule
\multicolumn{6}{l}{\textit{DrawerWorld}} \\
\midrule
Hold-Red-Box     & $\textbf{1538} \pm 130$  & $0 \pm 0$ &$0 \pm 0$          & $0\pm0$              & $0 \pm 0$       & $0 \pm 0$ \\
Pickup-Each-Box  & $\textbf{1.00} \pm 0.00$ & $0 \pm 0$ & $0 \pm 0$          & $0\pm0$              & $0 \pm 0$       & $0 \pm 0$ \\
Show-Green-Box   & $\textbf{0.61} \pm 0.06$ & $0 \pm 0$ & $0 \pm 0$          & $0\pm0$              & $0\pm0$         & $0\pm0$ \\
\bottomrule
\end{tabular}
\vspace{-1em}
\end{table}

\subsection{Extending Ground-Compose-Reinforce with a Natural Language Interface}

While natural language (NL) is often argued to have compositional properties \citep{sep-compositionality}, exploiting this compositionality in agentic language models (e.g. vision-language-action models) remains an open challenge. In Appendix~\ref{sec:autoformalization}, {we show that our GeoGrid RMs can be \emph{autoformalized} directly from an NL reward function description using OpenAI's o3 model}, \emph{zero-shot}---i.e., without fine-tuning on trajectories or other forms of grounding in our specific environments. Thus, we posit that leveraging compositional representations like RMs can be an effective way of building NL-interfaced agents in settings with limited labelled trajectory data (i.e. where $|\mathcal{D}|$ is small).


\section{Future Work and Limitations}
\label{sec:limitations}



\textbf{Extension to Other Compositional Representations:} In this work, we propose an end-to-end framework for grounding high-level specifications in behaviours that leverages the compositionality inherent in RMs. However, we believe our core insights apply to a wide range of compositional representations such as those that deal with objects \citep{diuk2008object} and relations \citep{dvzeroski2001relational}.


\textbf{Extension to Other Problem Settings:} Grounding language is a prerequisite for a myriad of {language-conditioned} problem settings. We consider an ``RL-in-the-loop'' setting, but future works could extend our insights to zero-shot execution of language tasks \citep{black2024pi_0, vaezipoor2021ltl2action, jackermeierdeepltl}, question answering \citep{hirschman2001natural, simmons1970natural}, and interactive task learning \citep{chai2018language, laird2017interactive}.


\textbf{Reward Hacking:} Misalignment between an agent’s interpretation of a task and human intent can lead to harmful consequences, particularly in RL \citep{skalse2022defining}. The use of formal specifications like RMs, which are unambiguous over the propositional vocabulary, can partially mitigate this, but ambiguity in the propositions themselves remains a concern in Ground-Compose-Reinforce. Prior works suggest that RM structure can be exploited to improve decision making under such ambiguity \citep{li2024reward, li2022noisy}. 


{\textbf{Assumptions on $\mathcal{D}$:} We assume that trajectories in $\mathcal{D}$ are labelled with values for a fixed set of propositions. Future works could explore other representations of propositions (e.g. as text) as well as scalable labelling methods (e.g. crowdsourced annotations \cite{stiennon2020learning} or self-supervised learning \cite{radford2021learning}).


\section{Conclusion}


This work presents Ground-Compose-Reinforce, an end-to-end framework for training RL agents directly from Reward Machine specifications---without oracle reward or labelling functions. A key challenge that we address is \emph{grounding} these high-level task specifications in executable behaviours, given an agent's perception and action capabilities. We find that exploiting compositional task structure is critical to faithfully capturing this grounding from limited data. Starting from only 350 labelled pretraining trajectories, we show that our technical approach scales to temporally extended manipulation tasks in Meta-World while generalizing out-of-distribution to behaviours that never appear in pretraining. Moreover, we show that in some cases, Reward Machines can be autoformalized directly from natural language reward function descriptions to expose this temporal task structure. 

More broadly, we show that leveraging language \emph{compositionality} presents a promising pathway to building language-driven agents \textit{without} relying on massive language-labelled data. Future work could explore the extension of these ideas to large-scale agentic language models such as vision-language-action models.

\section*{Acknowledgements}

We thank Harris Chan for his insightful and valuable input throughout all stages of this project. We gratefully acknowledge funding from the Natural Sciences and
Engineering Research Council of Canada (NSERC) and the Canada CIFAR AI Chairs
Program. Resources used in preparing this research
were provided, in part, by the Province of Ontario, the Government of
Canada through CIFAR, and companies sponsoring the Vector Institute for
Artificial Intelligence (\url{https://vectorinstitute.ai/partnerships/}). Finally, we
thank the Schwartz Reisman Institute for Technology and Society for
providing a rich multi-disciplinary research environment.
\bibliography{bib}

\providecommand{\Proceedings}{Proceedings\xspace} \providecommand{\International}{International\xspace} \providecommand{\Conference}{Conference\xspace} \providecommand{\Artificial}{Artificial\xspace} \providecommand{\Intelligence}{Intelligence\xspace} \providecommand{\AI}{Artificial Intelligence} \providecommand{\Scheduling}{Scheduling\xspace} \providecommand{\ofthe}{of the\xspace} \providecommand{\longshortnopar}[2]{#1} \providecommand{\longshort[2]}{#1 (#2)}
\begin{thebibliography}{88}
\providecommand{\natexlab}[1]{#1}
\providecommand{\url}[1]{\texttt{#1}}
\expandafter\ifx\csname urlstyle\endcsname\relax
  \providecommand{\doi}[1]{doi: #1}\else
  \providecommand{\doi}{doi: \begingroup \urlstyle{rm}\Url}\fi

\bibitem[Chevalier{-}Boisvert et~al.(2019)Chevalier{-}Boisvert, Bahdanau, Lahlou, Willems, Saharia, Nguyen, and Bengio]{chevalier2018babyai}
Maxime Chevalier{-}Boisvert, Dzmitry Bahdanau, Salem Lahlou, Lucas Willems, Chitwan Saharia, Thien~Huu Nguyen, and Yoshua Bengio.
\newblock Baby{AI}: {A} platform to study the sample efficiency of grounded language learning.
\newblock In \emph{\longshort{\Proceedings \ofthe 7th \International \Conference on Learning Representations}{ICLR}}, 2019.

\bibitem[Hermann et~al.(2017)Hermann, Hill, Green, Wang, Faulkner, Soyer, Szepesvari, Czarnecki, Jaderberg, Teplyashin, Wainwright, Apps, Hassabis, and Blunsom]{hermann2017grounded}
Karl~Moritz Hermann, Felix Hill, Simon Green, Fumin Wang, Ryan Faulkner, Hubert Soyer, David Szepesvari, Wojciech~Marian Czarnecki, Max Jaderberg, Denis Teplyashin, Marcus Wainwright, Chris Apps, Demis Hassabis, and Phil Blunsom.
\newblock Grounded language learning in a simulated 3d world.
\newblock \emph{arXiv preprint arXiv:1706.06551}, 2017.

\bibitem[Hill et~al.(2021)Hill, Tieleman, von Glehn, Wong, Merzic, and Clark]{hill2020grounded}
Felix Hill, Olivier Tieleman, Tamara von Glehn, Nathaniel Wong, Hamza Merzic, and Stephen Clark.
\newblock Grounded language learning fast and slow.
\newblock In \emph{\longshort{\Proceedings \ofthe 9th \International \Conference on Learning Representations}{ICLR}}, 2021.

\bibitem[Chaplot et~al.(2018)Chaplot, Sathyendra, Pasumarthi, Rajagopal, and Salakhutdinov]{chaplot2018gated}
Devendra~Singh Chaplot, Kanthashree~Mysore Sathyendra, Rama~Kumar Pasumarthi, Dheeraj Rajagopal, and Ruslan Salakhutdinov.
\newblock Gated-attention architectures for task-oriented language grounding.
\newblock In \emph{\longshort{\Proceedings \ofthe 32nd AAAI \Conference on \AI{}}{AAAI}}, volume~32, 2018.

\bibitem[Black et~al.(2024)Black, Brown, Driess, Esmail, Equi, Finn, Fusai, Groom, Hausman, Ichter, Jakubczak, Jones, Ke, Levine, Li{-}Bell, Mothukuri, Nair, Pertsch, et~al.]{black2024pi_0}
Kevin Black, Noah Brown, Danny Driess, Adnan Esmail, Michael Equi, Chelsea Finn, Niccolo Fusai, Lachy Groom, Karol Hausman, Brian Ichter, Szymon Jakubczak, Tim Jones, Liyiming Ke, Sergey Levine, Adrian Li{-}Bell, Mohith Mothukuri, Suraj Nair, Karl Pertsch, et~al.
\newblock $\pi_0 $: A vision-language-action flow model for general robot control.
\newblock \emph{arXiv preprint arXiv:2410.24164}, 2024.

\bibitem[Zitkovich et~al.(2023)Zitkovich, Yu, Xu, Xu, Xiao, Xia, Wu, Wohlhart, Welker, Wahid, Vuong, Vanhoucke, Tran, Soricut, Singh, Singh, Sermanet, Sanketi, Salazar, et~al.]{brohan2023rt}
Brianna Zitkovich, Tianhe Yu, Sichun Xu, Peng Xu, Ted Xiao, Fei Xia, Jialin Wu, Paul Wohlhart, Stefan Welker, Ayzaan Wahid, Quan Vuong, Vincent Vanhoucke, Huong Tran, Radu Soricut, Anikait Singh, Jaspiar Singh, Pierre Sermanet, Pannag~R. Sanketi, Grecia Salazar, et~al.
\newblock {RT}-2: Vision-language-action models transfer web knowledge to robotic control.
\newblock In \emph{\longshort{\Proceedings \ofthe 7th \Conference on Robot Learning}{CoRL}}, volume 229, pages 2165--2183. PMLR, 2023.

\bibitem[Ma et~al.(2023)Ma, Kumar, Zhang, Bastani, and Jayaraman]{ma2023liv}
Yecheng~Jason Ma, Vikash Kumar, Amy Zhang, Osbert Bastani, and Dinesh Jayaraman.
\newblock {LIV}: Language-image representations and rewards for robotic control.
\newblock In \emph{\longshort{\Proceedings \ofthe 40th \International \Conference on Machine Learning}{ICML}}, volume 202, pages 23301--23320. PMLR, 2023.

\bibitem[Baker et~al.(2022)Baker, Akkaya, Zhokov, Huizinga, Tang, Ecoffet, Houghton, Sampedro, and Clune]{baker2022video}
Bowen Baker, Ilge Akkaya, Peter Zhokov, Joost Huizinga, Jie Tang, Adrien Ecoffet, Brandon Houghton, Raul Sampedro, and Jeff Clune.
\newblock Video pretraining ({VPT}): Learning to act by watching unlabeled online videos.
\newblock In \emph{\longshort{\Proceedings \ofthe 36th \Conference on Advances in Neural Information Processing Systems}{NeurIPS}}, volume~35, pages 24639--24654, 2022.

\bibitem[Kaplan et~al.(2020)Kaplan, McCandlish, Henighan, Brown, Chess, Child, Gray, Radford, Wu, and Amodei]{kaplan2020scaling}
Jared Kaplan, Sam McCandlish, Tom Henighan, Tom~B Brown, Benjamin Chess, Rewon Child, Scott Gray, Alec Radford, Jeffrey Wu, and Dario Amodei.
\newblock Scaling laws for neural language models.
\newblock \emph{arXiv preprint arXiv:2001.08361}, 2020.

\bibitem[Aghajanyan et~al.(2023)Aghajanyan, Yu, Conneau, Hsu, Hambardzumyan, Zhang, Roller, Goyal, Levy, and Zettlemoyer]{aghajanyan2023scaling}
Armen Aghajanyan, Lili Yu, Alexis Conneau, Wei-Ning Hsu, Karen Hambardzumyan, Susan Zhang, Stephen Roller, Naman Goyal, Omer Levy, and Luke Zettlemoyer.
\newblock Scaling laws for generative mixed-modal language models.
\newblock In \emph{\longshort{\Proceedings \ofthe 40th \International \Conference on Machine Learning}{ICML}}, volume 202, pages 265--279. PMLR, 2023.

\bibitem[Stechly et~al.(2024)Stechly, Valmeekam, and Kambhampati]{StechlyNeurIPS2024thoughtlessness}
Kaya Stechly, Karthik Valmeekam, and Subbarao Kambhampati.
\newblock Chain of thoughtlessness? {A}n analysis of {CoT} in planning.
\newblock In \emph{\longshort{\Proceedings \ofthe 38th \Conference on Advances in Neural Information Processing Systems}{NeurIPS}}, volume~37, pages 29106--29141, 2024.

\bibitem[Shi et~al.(2025)Shi, Equi, Ke, Pertsch, Vuong, Tanner, Walling, Wang, Fusai, Li-Bell, et~al.]{shi2025hi}
Lucy~Xiaoyang Shi, Michael~Robert Equi, Liyiming Ke, Karl Pertsch, Quan Vuong, James Tanner, Anna Walling, Haohuan Wang, Niccolo Fusai, Adrian Li-Bell, et~al.
\newblock Hi robot: Open-ended instruction following with hierarchical vision-language-action models.
\newblock In \emph{\longshort{\Proceedings \ofthe 42nd \International \Conference on Machine Learning}{ICML}}, volume 267, pages 54919--54933. PMLR, 2025.

\bibitem[Lifshitz et~al.(2023)Lifshitz, Paster, Chan, Ba, and McIlraith]{lifshitz2023steve}
Shalev Lifshitz, Keiran Paster, Harris Chan, Jimmy Ba, and Sheila McIlraith.
\newblock {STEVE}-1: A generative model for text-to-behavior in {M}inecraft.
\newblock In \emph{\longshort{\Proceedings \ofthe 37th \Conference on Advances in Neural Information Processing Systems}{NeurIPS}}, volume~36, pages 69900--69929, 2023.

\bibitem[Bommasani et~al.(2021)Bommasani, Hudson, Adeli, Altman, Arora, von Arx, Bernstein, Bohg, Bosselut, Brunskill, Brynjolfsson, Buch, Card, Castellon, Chatterji, Chen, et~al.]{DBLP:journals/corr/abs-2108-07258}
Rishi Bommasani, Drew~A. Hudson, Ehsan Adeli, Russ~B. Altman, Simran Arora, Sydney von Arx, Michael~S. Bernstein, Jeannette Bohg, Antoine Bosselut, Emma Brunskill, Erik Brynjolfsson, Shyamal Buch, Dallas Card, Rodrigo Castellon, Niladri~S. Chatterji, Annie~S. Chen, et~al.
\newblock On the opportunities and risks of foundation models.
\newblock \emph{arXiv preprint arXiv:2108.07258}, 2021.

\bibitem[Toro~Icarte et~al.(2018)Toro~Icarte, Klassen, Valenzano, and McIlraith]{icarte2018using}
Rodrigo Toro~Icarte, Toryn Klassen, Richard Valenzano, and Sheila McIlraith.
\newblock Using reward machines for high-level task specification and decomposition in reinforcement learning.
\newblock In \emph{\longshort{\Proceedings \ofthe 35th \International \Conference on Machine Learning}{ICML}}, volume~80, pages 2107--2116. PMLR, 2018.

\bibitem[Toro~Icarte et~al.(2022)Toro~Icarte, Klassen, Valenzano, and McIlraith]{icarte2022reward}
Rodrigo Toro~Icarte, Toryn~Q Klassen, Richard Valenzano, and Sheila~A McIlraith.
\newblock Reward machines: Exploiting reward function structure in reinforcement learning.
\newblock \emph{Journal of Artificial Intelligence Research}, 73:\penalty0 173--208, 2022.

\bibitem[Yu et~al.(2020)Yu, Quillen, He, Julian, Hausman, Finn, and Levine]{yu2020meta}
Tianhe Yu, Deirdre Quillen, Zhanpeng He, Ryan Julian, Karol Hausman, Chelsea Finn, and Sergey Levine.
\newblock Meta-{W}orld: A benchmark and evaluation for multi-task and meta reinforcement learning.
\newblock In \emph{\longshort{\Proceedings \ofthe 3rd \Conference on Robot Learning}{CoRL}}, volume 100, pages 1094--1100. PMLR, 2020.

\bibitem[Liu et~al.(2023{\natexlab{a}})Liu, Suri, Mu, Zhou, and Finn]{liu2023simple}
Evan~Zheran Liu, Sahaana Suri, Tong Mu, Allan Zhou, and Chelsea Finn.
\newblock Simple embodied language learning as a byproduct of meta-reinforcement learning.
\newblock In \emph{\longshort{\Proceedings \ofthe 40th \International \Conference on Machine Learning}{ICML}}, volume 202, pages 21997--22008. PMLR, 2023{\natexlab{a}}.

\bibitem[Kim et~al.(2025)Kim, Pertsch, Karamcheti, Xiao, Balakrishna, Nair, Rafailov, Foster, Sanketi, Vuong, Kollar, Burchfiel, Tedrake, Sadigh, Levine, Liang, and Finn]{kim2024openvla}
Moo~Jin Kim, Karl Pertsch, Siddharth Karamcheti, Ted Xiao, Ashwin Balakrishna, Suraj Nair, Rafael Rafailov, Ethan~P Foster, Pannag~R Sanketi, Quan Vuong, Thomas Kollar, Benjamin Burchfiel, Russ Tedrake, Dorsa Sadigh, Sergey Levine, Percy Liang, and Chelsea Finn.
\newblock Open{VLA}: An open-source vision-language-action model.
\newblock In \emph{\longshort{\Proceedings \ofthe 8th \Conference on Robot Learning}{CoRL}}, volume 270, pages 2679--2713. PMLR, 2025.

\bibitem[Bahdanau et~al.(2019)Bahdanau, Hill, Leike, Hughes, Kohli, and Grefenstette]{bahdanau2018learning}
Dzmitry Bahdanau, Felix Hill, Jan Leike, Edward Hughes, Pushmeet Kohli, and Edward Grefenstette.
\newblock Learning to understand goal specifications by modelling reward.
\newblock In \emph{\longshort{\Proceedings \ofthe 7th \International \Conference on Learning Representations}{ICLR}}, 2019.

\bibitem[Baumli et~al.(2023)Baumli, Baveja, Behbahani, Chan, Comanici, Flennerhag, Gazeau, Holsheimer, Horgan, Laskin, Lyle, Masoom, McKinney, Mnih, Neitz, Pardo, et~al.]{baumli2023vision}
Kate Baumli, Satinder Baveja, Feryal M.~P. Behbahani, Harris Chan, Gheorghe Comanici, Sebastian Flennerhag, Maxime Gazeau, Kristian Holsheimer, Dan Horgan, Michael Laskin, Clare Lyle, Hussain Masoom, Kay McKinney, Volodymyr Mnih, Alexander Neitz, Fabio Pardo, et~al.
\newblock Vision-language models as a source of rewards.
\newblock \emph{arXiv preprint arXiv:2312.09187}, 2023.

\bibitem[Rocamonde et~al.(2024)Rocamonde, Montesinos, Nava, Perez, and Lindner]{rocamonde2023vision}
Juan Rocamonde, Victoriano Montesinos, Elvis Nava, Ethan Perez, and David Lindner.
\newblock Vision-language models are zero-shot reward models for reinforcement learning.
\newblock In \emph{\longshort{\Proceedings \ofthe 12th \International \Conference on Learning Representations}{ICLR}}, 2024.

\bibitem[Fu et~al.(2024)Fu, Zhang, Wu, Xu, and Boulet]{fu2024furl}
Yuwei Fu, Haichao Zhang, Di~Wu, Wei Xu, and Benoit Boulet.
\newblock {FuRL}: Visual-language models as fuzzy rewards for reinforcement learning.
\newblock In \emph{\longshort{\Proceedings \ofthe 41st \International \Conference on Machine Learning}{ICML}}, volume 235, pages 14256--14274. PMLR, 2024.

\bibitem[Yuan et~al.(2023)Yuan, Zhang, Wang, Xie, Cai, Dong, and Lu]{yuan2023plan4mc}
Haoqi Yuan, Chi Zhang, Hongcheng Wang, Feiyang Xie, Penglin Cai, Hao Dong, and Zongqing Lu.
\newblock {Plan4MC}: Skill reinforcement learning and planning for open-world {Minecraft} tasks.
\newblock \emph{arXiv preprint arXiv:2303.16563}, 2023.

\bibitem[Huang et~al.(2023)Huang, Xia, Xiao, Chan, Liang, Florence, Zeng, Tompson, Mordatch, Chebotar, Sermanet, Jackson, Brown, Luu, Levine, Hausman, and Ichter]{huang2022inner}
Wenlong Huang, Fei Xia, Ted Xiao, Harris Chan, Jacky Liang, Pete Florence, Andy Zeng, Jonathan Tompson, Igor Mordatch, Yevgen Chebotar, Pierre Sermanet, Tomas Jackson, Noah Brown, Linda Luu, Sergey Levine, Karol Hausman, and Brian Ichter.
\newblock Inner monologue: Embodied reasoning through planning with language models.
\newblock In \emph{\longshort{\Proceedings \ofthe 6th \Conference on Robot Learning}{CoRL}}, volume 205, pages 1769--1782. PMLR, 2023.

\bibitem[Ahn et~al.(2023)Ahn, Brohan, Brown, Chebotar, Cortes, David, Finn, Gopalakrishnan, Hausman, Herzog, Ho, Hsu, Ibarz, Ichter, Irpan, Jang, Ruano, et~al.]{ahn2022can}
Michael Ahn, Anthony Brohan, Noah Brown, Yevgen Chebotar, Omar Cortes, Byron David, Chelsea Finn, Keerthana Gopalakrishnan, Karol Hausman, Alexander Herzog, Daniel Ho, Jasmine Hsu, Julian Ibarz, Brian Ichter, Alex Irpan, Eric Jang, Rosario~Jauregui Ruano, et~al.
\newblock Do as {I} can, not as {I} say: Grounding language in robotic affordances.
\newblock In \emph{\longshort{\Proceedings \ofthe 6th \Conference on Robot Learning}{CoRL}}, volume 205, pages 287--318. PMLR, 2023.

\bibitem[Pnueli(1977)]{pnueli1977temporal}
Amir Pnueli.
\newblock The temporal logic of programs.
\newblock In \emph{18th Annual Symposium on Foundations of Computer Science}, pages 46--57. {IEEE}, 1977.

\bibitem[Karaman et~al.(2008)Karaman, Sanfelice, and Frazzoli]{karaman2008optimal}
Sertac Karaman, Ricardo~G. Sanfelice, and Emilio Frazzoli.
\newblock Optimal control of mixed logical dynamical systems with linear temporal logic specifications.
\newblock In \emph{Proceedings of the 47th {IEEE} Conference on Decision and Control, {CDC}}, pages 2117--2122. {IEEE}, 2008.

\bibitem[Kloetzer and Belta(2008)]{kloetzer2008fully}
Marius Kloetzer and Calin Belta.
\newblock A fully automated framework for control of linear systems from temporal logic specifications.
\newblock \emph{{IEEE} Transactions Automatic Control}, 53\penalty0 (1):\penalty0 287--297, 2008.

\bibitem[Baier and Katoen(2008)]{baier2008principles}
Christel Baier and Joost-Pieter Katoen.
\newblock \emph{Principles of Model Checking}.
\newblock MIT press, 2008.

\bibitem[Moon et~al.(1992)Moon, Powers, Burch, and Clarke]{moon1992automatic}
Il~Moon, Gary~J. Powers, Jerry~R. Burch, and Edmund~M. Clarke.
\newblock Automatic verification of sequential control systems using temporal logic.
\newblock \emph{AIChE Journal}, 38\penalty0 (1):\penalty0 67--75, 1992.

\bibitem[Pnueli(1986)]{pnueli2005applications}
Amir Pnueli.
\newblock Applications of temporal logic to the specification and verification of reactive systems: {A} survey of current trends.
\newblock In J.~W. de~Bakker, Willem~P. de~Roever, and Grzegorz Rozenberg, editors, \emph{Current Trends in Concurrency, Overviews and Tutorials}, volume 224 of \emph{Lecture Notes in Computer Science}, pages 510--584. Springer, 1986.

\bibitem[Li et~al.(2017)Li, Vasile, and Belta]{li2017reinforcement}
Xiao Li, Cristian~Ioan Vasile, and Calin Belta.
\newblock Reinforcement learning with temporal logic rewards.
\newblock In \emph{\longshort{\Proceedings \ofthe 2017 IEEE/RSJ \International \Conference on Intelligent Robots and Systems}{IROS}}, pages 3834--3839, 2017.

\bibitem[Camacho et~al.(2019)Camacho, Toro~Icarte, Klassen, Valenzano, and McIlraith]{camacho2019ltl}
Alberto Camacho, Rodrigo Toro~Icarte, Toryn~Q. Klassen, Richard Valenzano, and Sheila~A. McIlraith.
\newblock {LTL} and beyond: Formal languages for reward function specification in reinforcement learning.
\newblock In \emph{\longshort{\Proceedings \ofthe 28th \International Joint \Conference on \AI{}}{IJCAI}}, pages 6065--6073, 2019.

\bibitem[Voloshin et~al.(2023)Voloshin, Verma, and Yue]{voloshin2023eventual}
Cameron Voloshin, Abhinav Verma, and Yisong Yue.
\newblock Eventual discounting temporal logic counterfactual experience replay.
\newblock In \emph{\longshort{\Proceedings \ofthe 40th \International \Conference on Machine Learning}{ICML}}, volume 202, pages 35137--35150. PMLR, 2023.

\bibitem[Brunello et~al.(2019)Brunello, Montanari, and Reynolds]{brunello2019synthesis}
Andrea Brunello, Angelo Montanari, and Mark Reynolds.
\newblock Synthesis of {LTL} formulas from natural language texts: State of the art and research directions.
\newblock In \emph{26th International Symposium on Temporal Representation and Reasoning (TIME)}, volume 147 of \emph{LIPIcs}, pages 17:1--17:19. Schloss Dagstuhl--Leibniz-Zentrum f{\"u}r Informatik, 2019.

\bibitem[Liu et~al.(2023{\natexlab{b}})Liu, Yang, Idrees, Liang, Schornstein, Tellex, and Shah]{liu2022lang2ltl}
Jason~Xinyu Liu, Ziyi Yang, Ifrah Idrees, Sam Liang, Benjamin Schornstein, Stefanie Tellex, and Ankit Shah.
\newblock Grounding complex natural language commands for temporal tasks in unseen environments.
\newblock In \emph{\longshort{\Proceedings \ofthe 7th \Conference on Robot Learning}{CoRL}}, volume 229, pages 1084--1110. PMLR, 2023{\natexlab{b}}.

\bibitem[Fuggitti and Chakraborti(2023)]{fuggitti2023nl2ltl}
Francesco Fuggitti and Tathagata Chakraborti.
\newblock {NL2LTL}--a python package for converting natural language ({NL}) instructions to linear temporal logic ({LTL}) formulas.
\newblock In \emph{\longshort{\Proceedings \ofthe 37th AAAI \Conference on \AI{}}{AAAI}}, volume~37, pages 16428--16430, 2023.

\bibitem[Chen et~al.(2023)Chen, Gandhi, Zhang, and Fan]{chen2023nl2tl}
Yongchao Chen, Rujul Gandhi, Yang Zhang, and Chuchu Fan.
\newblock {NL}2{TL}: Transforming natural languages to temporal logics using large language models.
\newblock In \emph{Proceedings of the 2023 Conference on Empirical Methods in Natural Language Processing (EMNLP)}, pages 15880--15903, 2023.

\bibitem[Vaezipoor et~al.(2021)Vaezipoor, Li, Icarte, and Mcilraith]{vaezipoor2021ltl2action}
Pashootan Vaezipoor, Andrew~C Li, Rodrigo A~Toro Icarte, and Sheila~A Mcilraith.
\newblock {LTL2Action}: Generalizing {LTL} instructions for multi-task {RL}.
\newblock In \emph{\longshort{\Proceedings \ofthe 38th \International \Conference on Machine Learning}{ICML}}, volume 139, pages 10497--10508. PMLR, 2021.

\bibitem[Kuo et~al.(2020)Kuo, Katz, and Barbu]{kuo2020encoding}
Yen-Ling Kuo, Boris Katz, and Andrei Barbu.
\newblock Encoding formulas as deep networks: Reinforcement learning for zero-shot execution of {LTL} formulas.
\newblock In \emph{\longshort{\Proceedings \ofthe 2020 IEEE/RSJ \International \Conference on Intelligent Robots and Systems}{IROS}}, pages 5604--5610, 2020.

\bibitem[Yalcinkaya et~al.(2024)Yalcinkaya, Lauffer, Vazquez-Chanlatte, and Seshia]{NEURIPS2024_858fc542}
Beyazit Yalcinkaya, Niklas Lauffer, Marcell Vazquez-Chanlatte, and Sanjit~A. Seshia.
\newblock Compositional automata embeddings for goal-conditioned reinforcement learning.
\newblock In \emph{\longshort{\Proceedings \ofthe 38th \Conference on Advances in Neural Information Processing Systems}{NeurIPS}}, volume~37, pages 72933--72963, 2024.

\bibitem[Qiu et~al.(2023)Qiu, Mao, and Zhu]{NEURIPS2023_7b35a69f}
Wenjie Qiu, Wensen Mao, and He~Zhu.
\newblock Instructing goal-conditioned reinforcement learning agents with temporal logic objectives.
\newblock In \emph{\longshort{\Proceedings \ofthe 37th \Conference on Advances in Neural Information Processing Systems}{NeurIPS}}, volume~36, pages 39147--39175, 2023.

\bibitem[Liu et~al.(2024)Liu, Shah, Rosen, Jia, Konidaris, and Tellex]{liu2024skill}
Jason~Xinyu Liu, Ankit Shah, Eric Rosen, Mingxi Jia, George Konidaris, and Stefanie Tellex.
\newblock Skill transfer for temporal task specification.
\newblock In \emph{\longshort{\Proceedings \ofthe 2024 IEEE \International \Conference on Robotics and Automation}{ICRA}}, pages 2535--2541. IEEE, 2024.

\bibitem[Le{\'o}n et~al.(2022)Le{\'o}n, Shanahan, and Belardinelli]{leon2021nutshell}
Borja~G. Le{\'o}n, Murray Shanahan, and Francesco Belardinelli.
\newblock In a nutshell, the human asked for this: {L}atent goals for following temporal specifications.
\newblock In \emph{\longshort{\Proceedings \ofthe 10th \International \Conference on Learning Representations}{ICLR}}, 2022.

\bibitem[Jackermeier and Abate(2025)]{jackermeierdeepltl}
Mathias Jackermeier and Alessandro Abate.
\newblock {DeepLTL}: Learning to efficiently satisfy complex {LTL} specifications for multi-task {RL}.
\newblock In \emph{\longshort{\Proceedings \ofthe 13th \International \Conference on Learning Representations}{ICLR}}, 2025.

\bibitem[{Nangue Tasse} et~al.(2024){Nangue Tasse}, Jarvis, James, and Rosman]{tasse2024skill}
Geraud {Nangue Tasse}, Devon Jarvis, Steven James, and Benjamin Rosman.
\newblock Skill machines: Temporal logic skill composition in reinforcement learning.
\newblock In \emph{\longshort{\Proceedings \ofthe 12th \International \Conference on Learning Representations}{ICLR}}, 2024.

\bibitem[Furelos-Blanco et~al.(2021)Furelos-Blanco, Law, Jonsson, Broda, and Russo]{furelos2021induction}
Daniel Furelos-Blanco, Mark Law, Anders Jonsson, Krysia Broda, and Alessandra Russo.
\newblock Induction and exploitation of subgoal automata for reinforcement learning.
\newblock \emph{Journal of Artificial Intelligence Research}, 70:\penalty0 1031--1116, 2021.

\bibitem[Para\'{c} et~al.(2024)Para\'{c}, Nodari, Ardon, Furelos-Blanco, Cerutti, and Russo]{10.24963/kr.2024/85}
Roko Para\'{c}, Lorenzo Nodari, Leo Ardon, Daniel Furelos-Blanco, Federico Cerutti, and Alessandra Russo.
\newblock Learning robust reward machines from noisy labels.
\newblock In \emph{\longshort{\Proceedings \ofthe 21st \International \Conference on Knowledge Representation and Reasoning}{KR}}, 2024.

\bibitem[Jiang et~al.(2021)Jiang, Bharadwaj, Wu, Shah, Topcu, and Stone]{jiang2021temporal}
Yuqian Jiang, Suda Bharadwaj, Bo~Wu, Rishi Shah, Ufuk Topcu, and Peter Stone.
\newblock Temporal-logic-based reward shaping for continuing reinforcement learning tasks.
\newblock In \emph{\longshort{\Proceedings \ofthe 35th AAAI \Conference on \AI{}}{AAAI}}, volume~35, pages 7995--8003, 2021.

\bibitem[Elbarbari et~al.(2021)Elbarbari, Efthymiadis, Vanderborght, and Now{\'e}]{elbarbari2021ltlf}
Mahmoud Elbarbari, Kyriakos Efthymiadis, Bram Vanderborght, and Ann Now{\'e}.
\newblock Ltlf-based reward shaping for reinforcement learning.
\newblock In \emph{Adaptive and Learning Agents Workshop 2021: at AAMAS}, 2021.

\bibitem[Jothimurugan et~al.(2019)Jothimurugan, Alur, and Bastani]{jothimurugan2019composable}
Kishor Jothimurugan, Rajeev Alur, and Osbert Bastani.
\newblock A composable specification language for reinforcement learning tasks.
\newblock In \emph{\longshort{\Proceedings \ofthe 33rd \Conference on Advances in Neural Information Processing Systems}{NeurIPS}}, volume~32, 2019.

\bibitem[Balakrishnan and Deshmukh(2019)]{8968254}
Anand Balakrishnan and Jyotirmoy~V. Deshmukh.
\newblock Structured reward shaping using signal temporal logic specifications.
\newblock In \emph{\longshort{\Proceedings \ofthe 2019 IEEE/RSJ \International \Conference on Intelligent Robots and Systems}{IROS}}, pages 3481--3486, 2019.

\bibitem[Aksaray et~al.(2016)Aksaray, Jones, Kong, Schwager, and Belta]{7799279}
Derya Aksaray, Austin Jones, Zhaodan Kong, Mac Schwager, and Calin Belta.
\newblock Q-learning for robust satisfaction of signal temporal logic specifications.
\newblock In \emph{2016 IEEE 55th Conference on Decision and Control (CDC)}, pages 6565--6570, 2016.
\newblock \doi{10.1109/CDC.2016.7799279}.

\bibitem[Li et~al.(2024)Li, Chen, Klassen, Vaezipoor, Toro~Icarte, and McIlraith]{li2024reward}
Andrew Li, Zizhao Chen, Toryn Klassen, Pashootan Vaezipoor, Rodrigo Toro~Icarte, and Sheila McIlraith.
\newblock Reward machines for deep {RL} in noisy and uncertain environments.
\newblock In \emph{\longshort{\Proceedings \ofthe 38th \Conference on Advances in Neural Information Processing Systems}{NeurIPS}}, volume~37, pages 110341--110368, 2024.

\bibitem[Li et~al.(2022)Li, Chen, Vaezipoor, Klassen, Icarte, and McIlraith]{li2022noisy}
Andrew~C Li, Zizhao Chen, Pashootan Vaezipoor, Toryn~Q Klassen, Rodrigo~Toro Icarte, and Sheila~A McIlraith.
\newblock Noisy symbolic abstractions for deep {RL}: A case study with reward machines.
\newblock \emph{arXiv preprint arXiv:2211.10902}, 2022.

\bibitem[Hyde and Santos(2024)]{HydeECAI2024triggers}
Gregory Hyde and Eugene Santos, Jr.
\newblock Detecting hidden triggers: Mapping non-{M}arkov reward functions to {M}arkov.
\newblock In \emph{\longshort{\Proceedings \ofthe 27th European \Conference on \AI{}}{ECAI}}, volume 392, pages 1357--1364. {IOS} Press, 2024.

\bibitem[Christoffersen et~al.(2023)Christoffersen, Li, Icarte, and McIlraith]{christoffersen2023learning}
Phillip~JK Christoffersen, Andrew~C Li, Rodrigo~Toro Icarte, and Sheila~A McIlraith.
\newblock Learning symbolic representations for reinforcement learning of non-{M}arkovian behavior.
\newblock \emph{arXiv preprint arXiv:2301.02952}, 2023.

\bibitem[Umili et~al.(2024)Umili, Argenziano, and Capobianco]{umili2024neural}
Elena Umili, Francesco Argenziano, and Roberto Capobianco.
\newblock Neural reward machines.
\newblock In \emph{\longshort{\Proceedings \ofthe 27th European \Conference on \AI{}}{ECAI}}, pages 3055--3062. IOS Press, 2024.

\bibitem[Umili et~al.(2023)Umili, Capobianco, and De~Giacomo]{umili2023grounding}
Elena Umili, Roberto Capobianco, and Giuseppe De~Giacomo.
\newblock Grounding {LTL}f specifications in image sequences.
\newblock In \emph{\longshort{\Proceedings \ofthe 20th \International \Conference on Knowledge Representation and Reasoning}{KR}}, volume~19, pages 668--678, 2023.

\bibitem[Andreas et~al.(2017)Andreas, Klein, and Levine]{pmlr-v70-andreas17a}
Jacob Andreas, Dan Klein, and Sergey Levine.
\newblock Modular multitask reinforcement learning with policy sketches.
\newblock In \emph{\longshort{\Proceedings \ofthe 34th \International \Conference on Machine Learning}{ICML}}, volume~70, pages 166--175. PMLR, 2017.

\bibitem[Oh et~al.(2017)Oh, Singh, Lee, and Kohli]{pmlr-v70-oh17a}
Junhyuk Oh, Satinder Singh, Honglak Lee, and Pushmeet Kohli.
\newblock Zero-shot task generalization with multi-task deep reinforcement learning.
\newblock In \emph{\longshort{\Proceedings \ofthe 34th \International \Conference on Machine Learning}{ICML}}, volume~70, pages 2661--2670. PMLR, 2017.

\bibitem[Brafman and De~Giacomo(2019)]{brafman2019regular}
Ronen~I. Brafman and Giuseppe De~Giacomo.
\newblock Regular {D}ecision {P}rocesses: A model for non-{M}arkovian domains.
\newblock In \emph{\longshort{\Proceedings \ofthe 28th \International Joint \Conference on \AI{}}{IJCAI}}, pages 5516--5522, 2019.

\bibitem[De~Giacomo and Vardi(2013)]{de2013linear}
Giuseppe De~Giacomo and Moshe~Y Vardi.
\newblock Linear {T}emporal {L}ogic and {L}inear {D}ynamic {L}ogic on finite traces.
\newblock In \emph{\longshort{\Proceedings \ofthe 23rd \International Joint \Conference on \AI{}}{IJCAI}}, pages 854--860, 2013.

\bibitem[Baier and McIlraith(2006)]{baier2006planning}
Jorge~A Baier and Sheila~A McIlraith.
\newblock Planning with first-order temporally extended goals using heuristic search.
\newblock In \emph{\longshort{\Proceedings \ofthe 21st National \Conference on \AI{}}{AAAI}}, volume~21, pages 788--795, 2006.

\bibitem[Illanes et~al.(2019a)Illanes, Yan, {Toro Icarte}, and McIlraith]{illanesYTM2019symbolic}
Le\'on Illanes, Xi~Yan, Rodrigo {Toro Icarte}, and Sheila~A. McIlraith.
\newblock Symbolic planning and model-free reinforcement learning: Training taskable agents.
\newblock In \emph{\longshort{\Proceedings \ofthe 4th Multi-disciplinary \Conference on Reinforcement Learning and Decision}{RLDM}}, pages 191--195, 2019a.

\bibitem[Illanes et~al.(2020)Illanes, Yan, {Toro Icarte}, and McIlraith]{illanesYTM2020symbolic}
Le\'on Illanes, Xi~Yan, Rodrigo {Toro Icarte}, and Sheila~A. McIlraith.
\newblock Symbolic plans as high-level instructions for reinforcement learning.
\newblock In \emph{\longshort{\Proceedings \ofthe 30th \International \Conference on Automated Planning and \Scheduling}{ICAPS}}, volume~30, pages 540--550, 2020.

\bibitem[Clark and Amodei(2016)]{clark2016faulty}
Jack Clark and Dario Amodei.
\newblock Faulty reward functions in the wild, 2016.
\newblock URL \url{https://openai.com/index/faulty-reward-functions/}.
\newblock Blog post.

\bibitem[Stiennon et~al.(2020)Stiennon, Ouyang, Wu, Ziegler, Lowe, Voss, Radford, Amodei, and Christiano]{stiennon2020learning}
Nisan Stiennon, Long Ouyang, Jeffrey Wu, Daniel Ziegler, Ryan Lowe, Chelsea Voss, Alec Radford, Dario Amodei, and Paul~F Christiano.
\newblock Learning to summarize with human feedback.
\newblock In \emph{\longshort{\Proceedings \ofthe 34th \Conference on Advances in Neural Information Processing Systems}{NeurIPS}}, volume~33, pages 3008--3021, 2020.

\bibitem[Radford et~al.(2021)Radford, Kim, Hallacy, Ramesh, Goh, Agarwal, Sastry, Askell, Mishkin, Clark, Krueger, and Sutskever]{radford2021learning}
Alec Radford, Jong~Wook Kim, Chris Hallacy, Aditya Ramesh, Gabriel Goh, Sandhini Agarwal, Girish Sastry, Amanda Askell, Pamela Mishkin, Jack Clark, Gretchen Krueger, and Ilya Sutskever.
\newblock Learning transferable visual models from natural language supervision.
\newblock In \emph{\longshort{\Proceedings \ofthe 38th \International \Conference on Machine Learning}{ICML}}, volume 139, pages 8748--8763. PMLR, 2021.

\bibitem[Ng et~al.(1999)Ng, Harada, and Russell]{ng1999policy}
Andrew~Y Ng, Daishi Harada, and Stuart Russell.
\newblock Policy invariance under reward transformations: Theory and application to reward shaping.
\newblock In \emph{\longshort{\Proceedings \ofthe 16th \International \Conference on Machine Learning}{ICML}}, pages 278--287, 1999.

\bibitem[Goguen(1969)]{goguen1969logic}
J.~A. Goguen.
\newblock The logic of inexact concepts.
\newblock \emph{Synthese}, 19\penalty0 (3/4):\penalty0 325--373, 1969.

\bibitem[Schulman et~al.(2017)Schulman, Wolski, Dhariwal, Radford, and Klimov]{schulman2017proximal}
John Schulman, Filip Wolski, Prafulla Dhariwal, Alec Radford, and Oleg Klimov.
\newblock Proximal policy optimization algorithms.
\newblock \emph{arXiv preprint arXiv:1707.06347}, 2017.

\bibitem[Cho et~al.(2014)Cho, van Merri{\"e}nboer, Gulcehre, Bahdanau, Bougares, Schwenk, and Bengio]{cho2014learning}
Kyunghyun Cho, Bart van Merri{\"e}nboer, Caglar Gulcehre, Dzmitry Bahdanau, Fethi Bougares, Holger Schwenk, and Yoshua Bengio.
\newblock Learning phrase representations using {RNN} encoder{--}decoder for statistical machine translation.
\newblock In \emph{Proceedings of the 2014 Conference on Empirical Methods in Natural Language Processing ({EMNLP})}, pages 1724--1734, 2014.

\bibitem[Szabó(2024)]{sep-compositionality}
Zoltán~Gendler Szabó.
\newblock {Compositionality}.
\newblock In Edward~N. Zalta and Uri Nodelman, editors, \emph{The {Stanford} Encyclopedia of Philosophy}. Metaphysics Research Lab, Stanford University, 2024.

\bibitem[Diuk et~al.(2008)Diuk, Cohen, and Littman]{diuk2008object}
Carlos Diuk, Andre Cohen, and Michael~L Littman.
\newblock An object-oriented representation for efficient reinforcement learning.
\newblock In \emph{\longshort{\Proceedings \ofthe 25th \International \Conference on Machine Learning}{ICML}}, pages 240--247, 2008.

\bibitem[D{\v{z}}eroski et~al.(2001)D{\v{z}}eroski, De~Raedt, and Driessens]{dvzeroski2001relational}
Sa{\v{s}}o D{\v{z}}eroski, Luc De~Raedt, and Kurt Driessens.
\newblock Relational reinforcement learning.
\newblock \emph{Machine Learning}, 43:\penalty0 7--52, 2001.

\bibitem[Hirschman and Gaizauskas(2001)]{hirschman2001natural}
Lynette Hirschman and Robert Gaizauskas.
\newblock Natural language question answering: the view from here.
\newblock \emph{Natural Language Engineering}, 7\penalty0 (4):\penalty0 275--300, 2001.

\bibitem[Simmons(1970)]{simmons1970natural}
Robert~F Simmons.
\newblock Natural language question-answering systems: 1969.
\newblock \emph{Communications of the ACM}, 13\penalty0 (1):\penalty0 15--30, 1970.

\bibitem[Chai et~al.(2018)Chai, Gao, She, Yang, Saba-Sadiya, and Xu]{chai2018language}
Joyce~Y Chai, Qiaozi Gao, Lanbo She, Shaohua Yang, Sari Saba-Sadiya, and Guangyue Xu.
\newblock Language to action: Towards interactive task learning with physical agents.
\newblock In \emph{\longshort{\Proceedings \ofthe 27th \International Joint \Conference on \AI{}}{IJCAI}}, volume~7, pages 2--9, 2018.

\bibitem[Laird et~al.(2017)Laird, Gluck, Anderson, Forbus, Jenkins, Lebiere, Salvucci, Scheutz, Thomaz, Trafton, Wray, Mohan, and Kirk]{laird2017interactive}
John~E. Laird, Kevin~A. Gluck, John~R. Anderson, Kenneth~D. Forbus, Odest~Chadwicke Jenkins, Christian Lebiere, Dario~D. Salvucci, Matthias Scheutz, Andrea Thomaz, J.~Gregory Trafton, Robert~E. Wray, Shiwali Mohan, and James~R. Kirk.
\newblock Interactive task learning.
\newblock \emph{IEEE Intelligent Systems}, 32\penalty0 (4):\penalty0 6--21, 2017.

\bibitem[Skalse et~al.(2022)Skalse, Howe, Krasheninnikov, and Krueger]{skalse2022defining}
Joar Skalse, Nikolaus Howe, Dmitrii Krasheninnikov, and David Krueger.
\newblock Defining and characterizing reward hacking.
\newblock In \emph{\longshort{\Proceedings \ofthe 36th \Conference on Advances in Neural Information Processing Systems}{NeurIPS}}, volume~35, pages 9460--9471, 2022.

\bibitem[Sutton et~al.(1999)Sutton, Precup, and Singh]{sutton1999between}
Richard~S Sutton, Doina Precup, and Satinder Singh.
\newblock Between {MDP}s and semi-{MDP}s: A framework for temporal abstraction in reinforcement learning.
\newblock \emph{Artificial intelligence}, 112\penalty0 (1-2):\penalty0 181--211, 1999.

\bibitem[Nangue~Tasse et~al.(2020)Nangue~Tasse, James, and Rosman]{nangue2020boolean}
Geraud Nangue~Tasse, Steven James, and Benjamin Rosman.
\newblock A {B}oolean task algebra for reinforcement learning.
\newblock In \emph{\longshort{\Proceedings \ofthe 34th \Conference on Advances in Neural Information Processing Systems}{NeurIPS}}, volume~33, pages 9497--9507, 2020.

\bibitem[Ernst et~al.(2005)Ernst, Geurts, and Wehenkel]{ernst2005tree}
Damien Ernst, Pierre Geurts, and Louis Wehenkel.
\newblock Tree-based batch mode reinforcement learning.
\newblock \emph{Journal of Machine Learning Research}, 6:\penalty0 503--556, 2005.

\bibitem[Lyu et~al.(2022)Lyu, Ma, Li, and Lu]{lyu2022mildly}
Jiafei Lyu, Xiaoteng Ma, Xiu Li, and Zongqing Lu.
\newblock Mildly conservative {Q}-learning for offline reinforcement learning.
\newblock In \emph{\longshort{\Proceedings \ofthe 36th \Conference on Advances in Neural Information Processing Systems}{NeurIPS}}, volume~35, pages 1711--1724, 2022.

\bibitem[Vaswani et~al.(2017)Vaswani, Shazeer, Parmar, Uszkoreit, Jones, Gomez, Kaiser, and Polosukhin]{vaswani2017attention}
Ashish Vaswani, Noam Shazeer, Niki Parmar, Jakob Uszkoreit, Llion Jones, Aidan~N Gomez, {\L}ukasz Kaiser, and Illia Polosukhin.
\newblock Attention is all you need.
\newblock In \emph{\longshort{\Proceedings \ofthe 31st \Conference on Advances in Neural Information Processing Systems}{NIPS}}, volume~30, 2017.

\bibitem[Menghi et~al.(2021)Menghi, Tsigkanos, Pelliccione, Ghezzi, and Berger]{8859226}
Claudio Menghi, Christos Tsigkanos, Patrizio Pelliccione, Carlo Ghezzi, and Thorsten Berger.
\newblock Specification patterns for robotic missions.
\newblock \emph{IEEE Transactions on Software Engineering}, 47\penalty0 (10):\penalty0 2208--2224, 2021.

\end{thebibliography}

\newpage 
\appendix
\clearpage
\section*{NeurIPS Paper Checklist}


\begin{enumerate}

\item {\bf Claims}
    \item[] Question: Do the main claims made in the abstract and introduction accurately reflect the paper's contributions and scope?
    \item[] Answer: \answerYes{} 
    \item[] Justification: The claims in the abstract and introduction are reflected in the framework described in Section~\ref{sec:gcr}, the compositional reward shaping in Section~\ref{sec:sparse_rewards}, and the experimental results in Section~\ref{sec:experiments}.

\item {\bf Limitations}
    \item[] Question: Does the paper discuss the limitations of the work performed by the authors?
    \item[] Answer: \answerYes{} 
    \item[] Justification: Limitations are discussed in Section~\ref{sec:limitations}.

\item {\bf Theory assumptions and proofs}
    \item[] Question: For each theoretical result, does the paper provide the full set of assumptions and a complete (and correct) proof?
    \item[] Answer: \answerNA{} 
    \item[] Justification: The paper does not include theoretical results.

    \item {\bf Experimental result reproducibility}
    \item[] Question: Does the paper fully disclose all the information needed to reproduce the main experimental results of the paper to the extent that it affects the main claims and/or conclusions of the paper (regardless of whether the code and data are provided or not)?
    \item[] Answer: \answerYes{} 
    \item[] Justification: Major experimental details are reported in Section 7, with details like hyperparameters and network architectures in the appendix. 

\item {\bf Open access to data and code}
    \item[] Question: Does the paper provide open access to the data and code, with sufficient instructions to faithfully reproduce the main experimental results, as described in supplemental material?
    \item[] Answer: \answerYes{} 
    \item[] Justification: Code and data is released at \url{https://github.com/andrewli77/ground-compose-reinforce}.

\item {\bf Experimental setting/details}
    \item[] Question: Does the paper specify all the training and test details (e.g., data splits, hyperparameters, how they were chosen, type of optimizer, etc.) necessary to understand the results?
    \item[] Answer: \answerYes{} 
    \item[] Justification: See details in Appendix~\ref{sec:training_details}.

\item {\bf Experiment statistical significance}
    \item[] Question: Does the paper report error bars suitably and correctly defined or other appropriate information about the statistical significance of the experiments?
    \item[] Answer: \answerYes{} 
    \item[] Justification: Standard error is shown in Table~\ref{tab:results_rl}.

\item {\bf Experiments compute resources}
    \item[] Question: For each experiment, does the paper provide sufficient information on the computer resources (type of compute workers, memory, time of execution) needed to reproduce the experiments?
    \item[] Answer: \answerYes{} 
    \item[] Justification: See Appendix B.2 
    
\item {\bf Code of ethics}
    \item[] Question: Does the research conducted in the paper conform, in every respect, with the NeurIPS Code of Ethics \url{https://neurips.cc/public/EthicsGuidelines}?
    \item[] Answer: \answerYes{} 
    \item[] Justification: The code of ethics has been reviewed and the paper conforms with it.

\item {\bf Broader impacts}
    \item[] Question: Does the paper discuss both potential positive societal impacts and negative societal impacts of the work performed?
    \item[] Answer: \answerYes{} 
    \item[] Justification: Societal impacts are discussed in Appendix~\ref{sec:societalimpact}.
    
\item {\bf Safeguards}
    \item[] Question: Does the paper describe safeguards that have been put in place for responsible release of data or models that have a high risk for misuse (e.g., pretrained language models, image generators, or scraped datasets)?
    \item[] Answer: \answerNA{} 
    \item[] Justification: The data and models in this particular paper do not have any possible misuse potential. 

\item {\bf Licenses for existing assets}
    \item[] Question: Are the creators or original owners of assets (e.g., code, data, models), used in the paper, properly credited and are the license and terms of use explicitly mentioned and properly respected?
    \item[] Answer: \answerYes{} 
    \item[] Justification: Our environment is based on MetaWorld, which we cite. We also acknowledge the authors of the RL library we used. 

\item {\bf New assets}
    \item[] Question: Are new assets introduced in the paper well documented and is the documentation provided alongside the assets?
    \item[] Answer: \answerNA{} 
    \item[] Justification: No new assets are released at this time.

\item {\bf Crowdsourcing and research with human subjects}
    \item[] Question: For crowdsourcing experiments and research with human subjects, does the paper include the full text of instructions given to participants and screenshots, if applicable, as well as details about compensation (if any)? 
    \item[] Answer: \answerNA{} 
    \item[] Justification: The paper does not involve crowdsourcing nor research with human subjects.

\item {\bf Institutional review board (IRB) approvals or equivalent for research with human subjects}
    \item[] Question: Does the paper describe potential risks incurred by study participants, whether such risks were disclosed to the subjects, and whether Institutional Review Board (IRB) approvals (or an equivalent approval/review based on the requirements of your country or institution) were obtained?
    \item[] Answer: \answerNA{} 
    \item[] Justification: The paper does not involve crowdsourcing nor research with human subjects.

\item {\bf Declaration of LLM usage}
    \item[] Question: Does the paper describe the usage of LLMs if it is an important, original, or non-standard component of the core methods in this research? Note that if the LLM is used only for writing, editing, or formatting purposes and does not impact the core methodology, scientific rigorousness, or originality of the research, declaration is not required.
    \item[] Answer: \answerNA{} 
    \item[] Justification: The methods in this paper do not use LLMs.

\end{enumerate}
\newpage

\pagebreak

{\Large\bf Supplementary Material for \textit{Ground-Compose-Reinforce: Tasking Reinforcement Learning Agents through Formal Language}}

In this supplementary material, 
\begin{itemize}\item we provide further details and analysis on Ground-Compose-Reinforce  (Appendix~\ref{sec:composition_supp})
\item we provide experimental details (Appendix~\ref{sec:Experimental_Details}),
\item we show how we autoformalize natural language descriptions of reward functions into our Reward Machines with LLMs (Appendix~\ref{sec:autoformalization}), 
\item and we discuss the potential societal impact of this work (Appendix~\ref{sec:societalimpact}).
\end{itemize}

\section{Supplementary Details on Ground-Compose-Reinforce}
\label{sec:composition_supp}


In this section, we
\begin{itemize}
    \item discuss when Ground-Compose-Reinforce is likely to work well in practice (Appendix~\ref{subsec:gcr_analysis})
    \item elaborate on Section~\ref{subsec:logical_subtasks}'s description of how we approximate the optimal value function $V^*_\mathcal{R}(s,u)$, 
    \begin{itemize}\item first under the assumption that the RM does not contain self-loop transitions with non-zero rewards (Appendix~\ref{subsec:rm_vfs}),
    \item and then relaxing that assumption (Appendix~\ref{subsec:rm_vfs2});
    \end{itemize} 
    \item discuss approximation errors that can occur (Appendix~\ref{sec:approx_errors});
    \item describe how we train with offline RL the \emph{primitive value functions} used in the approximation (Appendix~\ref{sec:train_primitive});
    \item show how we use the approximate value functions for potential-based reward shaping (Appendix~\ref{sec:reward_shaping}).
\end{itemize}

\subsection{Analysis of Advantages and Assumptions}
\label{subsec:gcr_analysis}

For Ground-Compose-Reinforce to work well in practice, we assume that a faithful grounding of propositions can be learned during the pretraining phase from $\mathcal{D}$ (i.e. $\hat{\mathcal{L}}(s) \approx \mathcal{L}^*(s)$). If this is the case, the rewards generated by the core framework (Algorithm~\ref{alg:reinforce}) will faithfully capture tasks for any RM over propositions $\mathcal{AP}$, by design. To achieve this, $\mathcal{D}$ should provide sufficient coverage of the state space $\mathcal{S}$---but critically, it does not require sufficient coverage of the space of possible trajectories to generalize. Hence, Ground-Compose-Reinforce is able to reliably elicit desirable trajectories that are significantly out-of-distribution with respect to $\mathcal{D}$. 

Unlike many imitation-learning-based methods, Ground-Compose-Reinforce also does not rely on a high concentration of expert demonstrations in $\mathcal{D}$ (as evidenced by the fact that our agent reliably solves tasks, even when $\mathcal{D}$ contains only random-action trajectories). We can attribute this to the RL phase, where the agent fine-tunes its policy using its self-generated learning signal.

\subsection{Approximating $V^*_\mathcal{R}(s,u)$ (No Rewarding RM Self-Transitions)}
\label{subsec:rm_vfs}

To estimate $V^*_\mathcal{R}(s,u)$, we evaluate each outgoing transition $\langle u, u', \varphi, r \rangle$ from $u$ by combining the subtask value $V^*_{\Diamond \varphi}(s)$ with $v^*_\mathcal{R}(u')$.
While \citet{camacho2019ltl} treat RM transitions as singular actions, we treat RM transitions as temporally extended \emph{options} \citep{sutton1999between} that take a variable number of steps to execute. Specifically, for a transition $\langle u, u', \varphi, r \rangle$, we make the following assumptions about the corresponding option:

\begin{enumerate}[label=(\arabic*), itemsep=0.5em, topsep=0em, parsep=0pt]
\item It can be initiated if and only if the current RM state is $u$.
\item It optimizes for the subtask $\Diamond\varphi$ (i.e. satisfying $\varphi$ quickly and with high probability).
\item It either terminates when $\varphi$ is satisfied after a variable length of time $K$, or it never terminates.
\item The option will never result in a different transition in the RM than the one in question.
\end{enumerate}

\renewcommand{\algorithmicrequire}{\textbf{Input:}}
\renewcommand{\algorithmicensure}{\textbf{Output:}}

\begin{algorithm}
\caption{Value Iteration over RM States (Modified from \citet{camacho2019ltl}) }
\label{alg:original_rm_shaping}
\begin{algorithmic}[1]
\REQUIRE RM $\mathcal{R} = \langle \mathcal{U}, u_0, \mathcal{F}, \mathcal{AP}, \delta_u, \delta_r \rangle$, discount factor $\gamma_{\mathrm{RM}}$

\STATE Initialize $v^*_{\mathcal{R}}(u) \leftarrow 0$, $\forall u\in\mathcal{U}$
\WHILE{not converged}
    \FOR{$u$ in $\mathcal{U}$}
        \STATE $v^*_{\mathcal{R}}(u) \leftarrow \max_{\langle u', \varphi, r \rangle \in \delta(u)}{  \gamma_{\mathrm{RM}}(r + v^*_{\mathcal{R}}(u'))  }$
    \ENDFOR
\ENDWHILE
\RETURN $v^*_{\mathcal{R}}$
\end{algorithmic}
\end{algorithm}

To estimate $v^*_\mathcal{R}(u)$, we run a variant of Value Iteration, modified from \citet{camacho2019ltl} to reflect that rewards are garnered \emph{after} the option completes (Algorithm~\ref{alg:original_rm_shaping}). Note that Algorithm~\ref{alg:original_rm_shaping} should be run with a high-level discount factor $\gamma_{\mathrm{RM}} \ll \gamma$ as it treats RM transitions as singular actions while in the actual RM-MDP, an RM transition may require many low-level steps to achieve.

We estimate the return for an RM transition $\langle u, u', \varphi, r \rangle$, given that the agent is in MDP state $s$ and RM state $u$, as follows. First, we model the return for \emph{immediately} achieving the transition as $r + \gamma v^*_\mathcal{R}(u')$ --- reward $r$ is immediately received for achieving the RM transition and $v^*_\mathcal{R}(u')$ estimates the future discounted return for being in RM state $u'$. If instead the RM transition is achieved $k$ steps in the future, we estimate the discounted return as $\gamma^k(r + \gamma v^*_\mathcal{R}(u'))$. Treating $\langle u, u', \varphi, r \rangle$ as an option that terminates when the RM transition is achieved, define $K$ to be the (random variable) number of steps it takes for this option to terminate when initiated from MDP state $s$ and RM state $u$ (where $K$ is $\infty$ if the RM transition is never achieved). The expected return of initiating the option is:

\begin{align*}
\mathbb{E}_{k \sim K} \bigl[\gamma^k(r + \gamma v^*_\mathcal{R}(u')) \bigr] &= (\mathbb{E}_{k \sim K} [\gamma^k])(r + \gamma v^*_\mathcal{R}(u')) \\
&= V^*_{\Diamond \varphi}(s) \cdot (r + \gamma v^*_\mathcal{R}(u'))
\end{align*}

Recalling that the option is assumed to optimize for the subtask $\Diamond \varphi$, the final step results from the fact that $\mathbb{E}_{k \sim K}[\gamma^k]$ is precisely the optimal value for $\Diamond \varphi$ from state $s$. We finally estimate the optimal value function for RM task $\mathcal{R}$ by maximizing over the choice of option: 
\begin{align*}
{V}^*_{\mathcal{R}}(s,u) &\approx \max_{\langle u', \varphi, r \rangle \in \delta(u)}{\bigl[ V^*_{\Diamond \varphi}(s) (r + \gamma v^*_\mathcal{R}(u'))\bigr]}
\end{align*}



\subsection{Approximating $V^*_\mathcal{R}(s,u)$ with Rewarding Self-Loop Transitions}
\label{subsec:rm_vfs2}

To handle RMs that contain self-loop transitions with non-zero rewards, we require a few modifications. Intuitively, an option corresponding to a transition $\langle u, u', \varphi, r \rangle$ where $u \neq u'$ might receive a reward $r'$ at any timestep it is active if there exists another transition $\langle u,u,\varphi',r' \rangle$. As a simple example, consider the RM in Figure~\ref{fig:rm_loop_example}. The RM describes a task where the agent needs to exit the lava as quickly as possible, and receives a reward of $-1$ for each step until it does so. When estimating the expected return of the edge labelled $\langle \lnot \textcolor{red}{lava},0 \rangle$, we need to consider the accumulation of the $-1$ reward at each timestep.  

\begin{figure}
    \centering
    \includegraphics[width=0.3\linewidth]{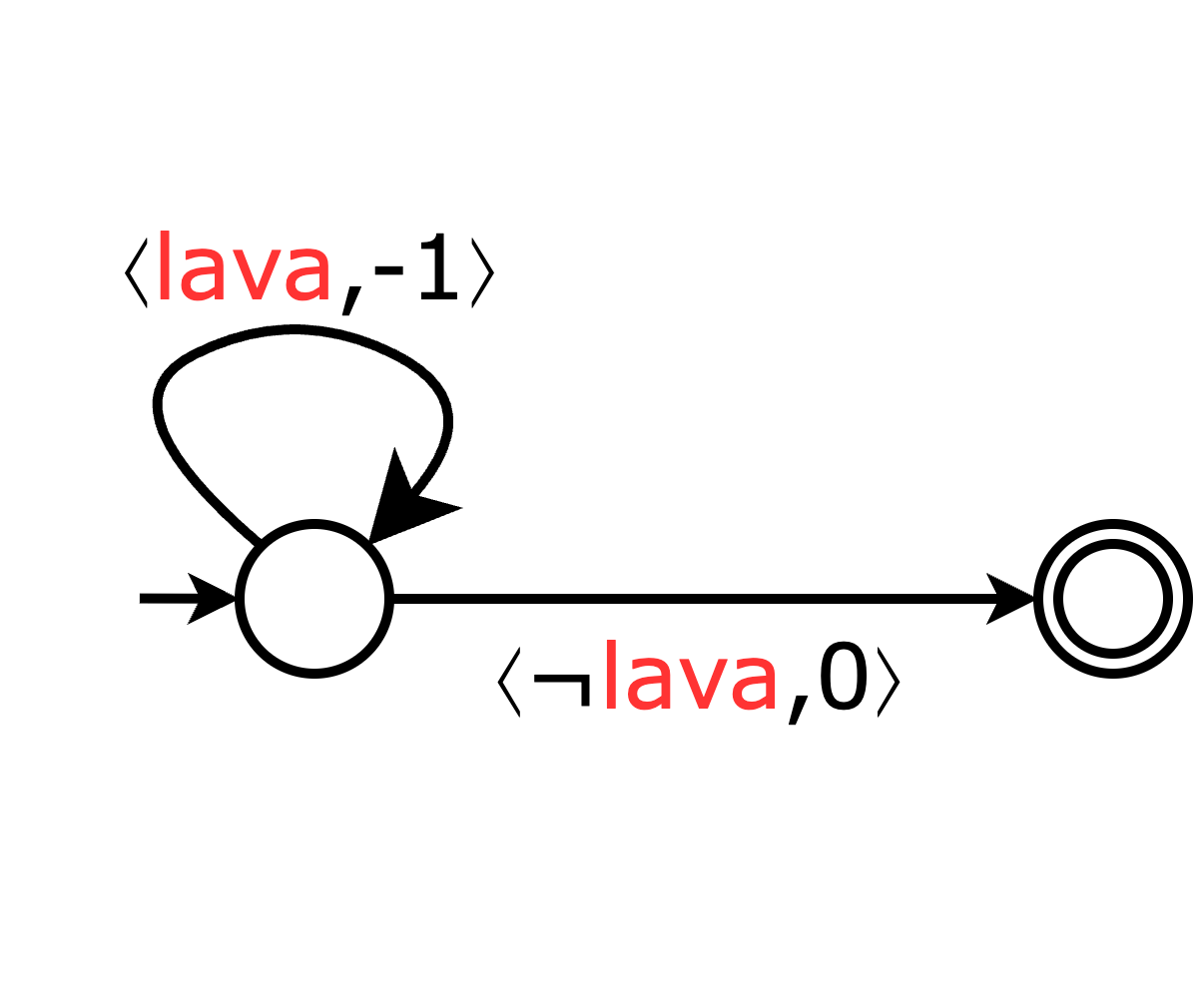}
    \vspace{-3em}
    \caption{An RM that produces a reward of $-1$ for each timestep the agent spends in lava, until it exits the lava.}
    \label{fig:rm_loop_example}
\end{figure}

This can be hard to handle in general without more information, particularly when multiple self-loop edges exist for the same RM state. For simplicity, we assume that only non-self-loop transitions are treated as options (for the purposes of estimating optimal values), while \emph{all} self-loop transitions in the current state $u$ are available at each step while an option is being executed. Thus, if we define $r_{u,u}$ as the \emph{maximum} reward for any self-loop transition in RM state $u$, we can modify our expression for the expected return of an option corresponding to transition $\langle u,u',\varphi,r \rangle$ in state $\langle s, u \rangle$ as follows:
\begin{align*}
&\mathbb{E}_{k \sim K, s'} \bigl[\textcolor{red}{r_{u,u} + \gamma r_{u,u} + \ldots + \gamma^{k-1}r_{u,u}} + \gamma^k(r + \gamma V^*_\mathcal{R}(s',u')) \bigr] \\ 
&= \mathbb{E}_{k \sim K, s'} \Bigl[\textcolor{red}{r_{u,u}\Bigl(\frac{1-\gamma^k}{1-\gamma}\Bigr)} + \gamma^k(r + \gamma V^*_\mathcal{R}(s',u')) \Bigr] \\
&= \mathbb{E}_{s'} \Bigl[ \textcolor{red}{r_{u,u}\Bigl(\frac{1-V^*_{\Diamond \varphi}(s)}{1-\gamma}\Bigr)} +  V^*_{\Diamond \varphi}(s) (r + \gamma V^*_\mathcal{R}(s',u')) \Bigr]  \\
&\approx \textcolor{red}{r_{u,u}\Bigl(\frac{1-V^*_{\Diamond \varphi}(s)}{1-\gamma}\Bigr)} +  V^*_{\Diamond \varphi}(s) (r + \gamma v^*_\mathcal{R}(u'))
\end{align*}
For the previous lava example, the expected return for the option that corresponds to the transition labelled $\langle \lnot \textcolor{red}{lava},0 \rangle$ now correctly reflects a reward of $-1$ obtained for each step until the option terminates by the agent exiting the lava. We modify our approximation of ${V}^*_{\mathcal{R}}(s,u)$ as follows:
\begin{align}
{V}^*_{\mathcal{R}}(s,u)
    &\approx \max_{\langle u', \varphi, r \rangle \in \delta(u)}{ \Bigl [  \textcolor{red}{r_{u,u}\Bigl(\frac{1-V^*_{\Diamond \varphi}(s)}{1-\gamma}\Bigr)} + V^*_{\Diamond \varphi}(s) (r + \gamma v^*_\mathcal{R}(u')) \Bigr ] }
\label{eq:value_rm_modified}
\end{align}

We also modify the Value Iteration algorithm to consider self-loops with non-zero rewards (see Algorithm~\ref{alg:modified_high_rm_shaping}).

\renewcommand{\algorithmicrequire}{\textbf{Input:}}
\renewcommand{\algorithmicensure}{\textbf{Output:}}

\begin{algorithm}
\caption{Value Iteration over RM States (modified for self-loop transitions) }
\label{alg:modified_high_rm_shaping}
\begin{algorithmic}[1]
\REQUIRE RM $\mathcal{R} = \langle \mathcal{U}, u_0, \mathcal{F}, \mathcal{AP}, \delta_u, \delta_r \rangle$, \textcolor{red}{MDP discount factor $\gamma$}, high-level discount factor $\gamma_{\mathrm{RM}}$

\STATE Initialize $v^*_{\mathcal{R}}(u) \leftarrow 0$, $\forall u\in\mathcal{U}$
\STATE \textcolor{red}{Extract maximum self-loop rewards $r_{u,u}$, $\forall u \in \mathcal{U}$ from $\delta_r$} 
\WHILE{not converged}
    \FOR{$u$ in $\mathcal{U}$}
        \STATE $v^*_{\mathcal{R}}(u) \leftarrow \max_{\langle u', \varphi, r \rangle \in \delta(u)}{ \bigl(\textcolor{red}{r_{u,u} \bigl( \frac{1 - \gamma_{\mathrm{RM}}}{\gamma} \bigr)} + \gamma_{\mathrm{RM}}(r + v^*_{\mathcal{R}}(u')) \bigr)}$
    \ENDFOR
\ENDWHILE
\RETURN $v^*_{\mathcal{R}}$
\end{algorithmic}
\end{algorithm}

\subsection{Sources of Approximation Errors}
\label{sec:approx_errors}

We now discuss how errors can occur when estimating $V^*_\mathcal{R}(s,u)$ from PVFs. 

First, Approximation~\ref{eq:value_rm_modified} clearly introduces approximation error from assuming the largest reward among self-loop transitions $r_{u,u}$ will be garnered until an outgoing transition is reached, and from using a bootstrapped value estimate of the next state $v^*_{\mathcal{R}}(u') \approx V^*_{\mathcal{R}}(s',u')$. Another less obvious source of error is that the expected return for each outgoing transition $\langle u, u', \varphi, r\rangle$ is estimated by assuming that an optimal policy for reaching $\varphi$ will be followed. However, this ignores the possibility of some \emph{other} transition from $u$ occurring before the intended transition. In general, it may not be possible to satisfy $\varphi$ while achieving value $V^*_{\Diamond \varphi}(s)$ in the subtask $\Diamond \varphi$ while also avoiding all other transitions from RM state $u$ that are not $\langle u, u', \varphi, r\rangle$. 

Estimating the OVF for a disjunction of formulas via Approximation~\ref{eq:disjunction} always \emph{underestimates} the true value. We prove this as follows. Suppose $\varphi = \xi_1 \lor \ldots \lor \xi_k$ and recall that the approximation of $V^*_{\Diamond \varphi}(s)$ is $\max_{i=1,\ldots,k} V^*_{\Diamond \xi_i}(s)$. Observe that for each $i$, $V^*_{\Diamond \xi_i}(s) \le V^*_{\Diamond \varphi}(s)$, since for every trajectory $\tau$, if $\xi_i$ is satisfied at timestep $T$ in $\tau$, then $\varphi$ must also be satisfied at timestep $T$ (or earlier) in $\tau$. Thus, $\max_{i=1,\ldots,k} V^*_{\Diamond \xi_i}(s) \le V^*_{\Diamond \varphi}(s)$. The reason this bound is not tight is because there are situations where satisfying \emph{one of} $\xi_1, \ldots, \xi_k$ is easier than satisfying any of $\xi_1, \ldots, \xi_k$ individually. For example, consider the task $\Diamond (X \lor \lnot X)$, which is always trivially solved on the first step. However, it is possible that $\Diamond X$ and $\Diamond \lnot X$ are both non-trivial tasks. 

Lastly, estimating the OVF for a conjunction of formulas via Approximation~\ref{eq:conjunction} always \emph{overestimates} the true value, by a similar line of reasoning as for disjunction. In general, knowing $V^*_{\Diamond \xi_1}(s)$ and $V^*_{\Diamond \xi_2}(s)$ does not provide enough information to estimate $V^*_{\Diamond (\xi_1 \land \xi_2)}(s)$. For instance, it may be the case that $\xi_1, \xi_2$ are mutually exclusive and thus,  $V^*_{\Diamond (\xi_1 \land \xi_2)}$ is zero everywhere. However, whether or not $\xi_1, \xi_2$ are mutually exclusive cannot be inferred based only on $V^*_{\Diamond \xi_1}(s)$ and $V^*_{\Diamond \xi_2}(s)$. \citet{nangue2020boolean} provide a discussion on this topic for a related setting.




\subsection{Training Primitive Value Functions}
\label{sec:train_primitive}

PVFs can be trained directly from the trajectory dataset $\mathcal{D}$ based on any offline RL approach. Algorithm~\ref{alg:grounding} shows how to train the PVF for a single primitive task $\Diamond x$ using a simple offline RL algorithm (Fitted Q-Iteration  \cite{ernst2005tree}). The approach can be easily adapted to negations $\Diamond \lnot x$ as well, and in practice, we simultaneously train all $2|\mathcal{AP}|$ possible PVFs in parallel as a single neural network.

\renewcommand{\algorithmicrequire}{\textbf{Input:}}
\renewcommand{\algorithmicensure}{\textbf{Output:}}

\begin{algorithm}[h]
\caption{Learning PVF $V^*_{\Diamond_{x}}(s)$ for $x \in \mathcal{AP}$ from Trajectory Data $\mathcal{D}$}
\label{alg:grounding}
\begin{algorithmic}[1]
\REQUIRE Dataset $\mathcal{D} = \left\{\langle \tau^i, \omega^i \rangle \right\}_{i=1}^{N}$, discount factor $\gamma$, proposition $x \in \mathcal{AP}$
\STATE Initialize Q function $Q_\theta: \mathcal{S} \times \mathcal{A} \to \mathbbm{R}$
\STATE Initialize value function $V_\psi: \mathcal{S} \to \mathbbm{R}$
\WHILE{not converged}
\STATE Sample transition $ \langle s,\omega, a,s',\omega' \rangle \sim \mathcal{D}$
\STATE Update $\phi$ with SGD on 
$\mathrm{BinaryCrossEntropy}(\mathcal{L}_\phi(s), \mathbbm{1}[x \in \omega])$ 
\STATE $\mathrm{reward} \gets \mathbbm{1} \left[ x \in \omega' \right], \mathrm{next\_value} \gets \max_{a' \in \mathcal{A}}Q_\theta(s', a'), \mathrm{done} \gets \mathbbm{1}\left[ x \in \omega' \right]$
\STATE Update $\theta$ with SGD on $\bigl(Q_\theta(s,a) - \mathrm{stop\_grad}(\mathrm{reward} + \gamma * (1-\mathrm{done}) * \mathrm{next\_value}) \bigr)^2$
\STATE Update $\psi$ with SGD on $(V_\psi(s') - \mathrm{next\_value})^2$
\ENDWHILE
\RETURN $V_\psi$
\end{algorithmic}
\end{algorithm}

\subsection{Potential-based Reward Shaping}
\label{sec:reward_shaping}

We extend our core Ground-Compose-Reinforce algorithm with potential-based reward shaping by leveraging Approximations~\ref{eq:value_rm}-\ref{eq:conjunction} and trained PVFs to predict $V^*_{\mathcal{R}}(s,u)$ for any MDP state $s$ and RM state $u$. This is shown in Algorithm~\ref{alg:reinforce2} with changes from the core algorithm highlighted in red.

\renewcommand{\algorithmicrequire}{\textbf{Input:}}
\renewcommand{\algorithmicensure}{\textbf{Output:}}

\begin{algorithm}[h]
\caption{Ground-Compose-Reinforce for RMs \textcolor{red}{with Potential-Based Reward Shaping}}
\label{alg:reinforce2}
\begin{algorithmic}[1]
\small
\REQUIRE MDP $\mathcal{M}$ without rewards, Propositional symbols $\mathcal{AP}$, Dataset $\mathcal{D}$ of labelled trajectories, RM task $\mathcal{R}$ over $\mathcal{AP}$, \textcolor{red}{Shaping potential weighting coefficient $\lambda$}

\COMMENT{Pretraining phase}
\STATE Train labelling function $\hat{\mathcal{L}}(s)$ on $\mathcal{D}$ using any binary classification method
\STATE \textcolor{red}{Train PVFs $V^*_{\Diamond x}(s)$ and  $V^*_{\Diamond \lnot x}(s)$, $\forall x\in \mathcal{AP}$ on $\mathcal{D}$}

\COMMENT{Behaviour elicitation phase}
\STATE Initialize policy $\pi_{\mathcal{R}}(a \mid s, u)$ arbitrarily
\FOR{each episode}
    \STATE Observe initial state $s$ in $\mathcal{M}$; set $u$ to the initial state of $\mathcal{R}$
    \STATE \textcolor{red}{Estimate initial value $v \approx V^*_{\mathcal{R}}(s,u)$ using Approximations~\ref{eq:value_rm}-\ref{eq:conjunction} and trained PVFs}
    \WHILE{$u$ is non-terminal}
        \STATE Sample action $a \sim \pi_{\mathcal{R}}(\cdot \mid s, u)$
        \STATE Execute $a$ in $\mathcal{M}$ and observe next state $s'$
        \STATE Compute truth assignment $\hat{\omega} \gets \hat{\mathcal{L}}(s')$
        \STATE Update RM: $u' \gets \delta_u(u, \hat{\omega})$, $r \gets \delta_r(u, \hat{\omega})$
        \STATE \textcolor{red}{Update value $v' \approx V^*_{\mathcal{R}}(s',u')$ using Approximations~\ref{eq:value_rm}-\ref{eq:conjunction} and trained PVFs}
        \STATE Update policy $\pi_{\mathcal{R}}$ with RL for transition $\langle s, u, a, r + \textcolor{red}{\lambda(\gamma v' - v)}, s', u' \rangle$
        \STATE Set $s \gets s'$, $u \gets u', \textcolor{red}{v \gets v'}$
    \ENDWHILE
\ENDFOR
\end{algorithmic}
\end{algorithm}

\section{Experimental Details}
\label{sec:Experimental_Details}

\subsection{Domain Descriptions}
\label{sec:domain_details}

\paragraph{Environments.} \textit{GeoGrid} is an $8 \times 8$ image-based gridworld depicted in Figure~\ref{fig:rm_examples} with six objects randomly positioned at the start of each episode. States are $8 \times 8 \times 6$-dimensional images that identify each cell's colour/shape and the agent's location, while propositions $\{\mathrm{\textcolor{red}{R}}, \mathrm{\textcolor{green}G}, \mathrm{\textcolor{blue}B}, \triangle, \Circle \}$ identify if the agent is at an object with that particular colour or shape. \emph{DrawerWorld} is a MuJoCo environment adapted from Meta-World \citep{yu2020meta}. The agent controls a robotic gripper and can interact with two drawers (left and right) and three boxes (red, green, and blue). Observations are 78-dimensional vectors representing positions of objects and the gripper. Propositions identify whether a particular drawer is open, whether a particular block is picked up by the agent, and whether a particular block is currently inside a particular drawer. 

\paragraph{Datasets.} We carefully curated datasets $\mathcal{D}$ in each environment to support our analysis of compositional generalization (RQ2).
In GeoGrid, $\mathcal{D}$ is comprised of 5000 trajectories of length 100 \emph{generated under a random-action policy}. In DrawerWorld, $\mathcal{D}$ is comprised of 350 trajectories of varying length that we collected by manually controlling the robot in the simulator. To ensure sufficient state coverage in $\mathcal{D}$, drawers and boxes were initialized in a random configuration when collecting each trajectory. Behaviours that appear in the dataset include opening and closing drawers, picking up boxes, and moving boxes from one location to another, but \emph{no trajectory involves direct interaction with more than one box}. A small number of trajectories involve incidental (but not prolonged) interaction with more than one box (e.g. bumping into one box while moving another). We intentionally include accidental behaviours in the DrawerWorld dataset such as failing to grip a box, dropping a box while attempting to move it, and opening a drawer beyond its limit. We also include behaviours not tied to downstream tasks such as placing a box on top of a drawer or throwing a box off the table.


\paragraph{Tasks.} We designed a diverse set of RM tasks (Table~\ref{tab:tasks}). For tasks that involve achieving a (temporally extended) goal, the RM terminates and provides a reward of 1 upon doing so. The RM terminates with a reward of 0 if the goal becomes logically impossible (e.g. due to breaking a constraint), except in Safety, which terminates with a penalty of $-1$. Loop and Hold-Red-Box involve repeating some desired behaviour, and the RM yields a reward of 1 for each such repetition. For the precise encodings of tasks as RMs, please see the released code.

\subsection{Baseline Implementation Details}

\begin{table}[t]
\scriptsize
\centering
\caption{Network Architectures for Supervised Training on $\mathcal{D}$.}
\label{tab:networks1}
\setlength{\tabcolsep}{2pt}
\begin{tabular}{ccccc}
\toprule
\multicolumn{2}{c}{\textbf{GCR}} & 
\textbf{LTL-BC} & 
\textbf{Bespoke Reward Model} & 
\textbf{Bespoke BC} \\
\textbf{Labelling Function} & \textbf{PVFs} & & & \\
\midrule
\multicolumn{5}{l}{\textit{GeoGrid}} \\
\midrule
\makecell[l]{
\texttt{Conv2d(6,16,3,1,1)} \\
\texttt{ReLU} \\
\texttt{Conv2d(16,32,3,1,1)} \\
\texttt{ReLU} \\
\texttt{Flatten} \\
\texttt{Linear(2048,128)} \\
\texttt{ReLU} \\
\texttt{Linear(128,5)}
} &

\makecell[l]{
\texttt{Conv2d(6,32,3,1,1)} \\
\texttt{ReLU} \\
\texttt{Conv2d(32,32,3,1,1)} \\
\texttt{ReLU} \\
\texttt{Flatten} \\
\texttt{Linear(2048,256)} \\
\texttt{ReLU} \\
\texttt{Linear(256,10)}
} &

\makecell[l]{
\textbf{Obs Encoder:} \\
\texttt{Conv2d(6,16,3,1,1)} \\
\texttt{ReLU} \\
\texttt{Conv2d(16,32,3,1,1)} \\
\texttt{ReLU} \\
\texttt{Flatten} \\
\texttt{Linear(2048,256)} \\
\\
\textbf{LTL Encoder:} \\
\texttt{Transformer(d\_model=64,} \\
\texttt{~~~nhead=4,} \\ 
\texttt{~~~dim\_feedforward=128, } \\
\texttt{~~~num\_layers=2)} \\ 
\\
\textbf{Policy:} \\
\texttt{GRU(256+64,256)} \\
\texttt{ReLU} \\
\texttt{GRU(256,256)} \\
\texttt{Linear(768,16)}
} &

\makecell[l]{
\texttt{Conv2d(6,16,3,1,1)} \\
\texttt{ReLU} \\
\texttt{Conv2d(16,32,3,1,1)} \\
\texttt{ReLU} \\
\texttt{Flatten} \\
\texttt{Linear(2048,256)} \\
\texttt{GRU(256,256)} \\
\texttt{ReLU} \\
\texttt{GRU(256,256)} \\
\texttt{Linear(768,4) $\times$ 3}
} &

\makecell[l]{
\texttt{Conv2d(6,16,3,1,1)} \\
\texttt{ReLU} \\
\texttt{Conv2d(16,32,3,1,1)} \\
\texttt{ReLU} \\
\texttt{Flatten} \\
\texttt{Linear(2048,256)} \\
\texttt{GRU(256,256)} \\
\texttt{ReLU} \\
\texttt{GRU(256,256)} \\
\texttt{Linear(768,16)}
} \\
\midrule
\multicolumn{5}{l}{\textit{DrawerWorld}} \\
\midrule
\makecell[l]{
\texttt{Linear(39,1600)} \\
\texttt{ReLU} \\
\texttt{Linear(1600,11)}
} &

\makecell[l]{
\texttt{Linear(39,1600)} \\
\texttt{ReLU} \\
\texttt{Linear(1600,400)} \\
\texttt{ReLU} \\
\texttt{Linear(400,22)}
} &

\makecell[l]{
\textbf{Obs Encoder:} \\
\texttt{Linear(39,1600)} \\
\texttt{ReLU} \\
\texttt{Linear(1600,400)} \\
\\
\textbf{LTL Encoder:} \\
\texttt{Transformer(d\_model=64,} \\
\texttt{~~~nhead=4,} \\ 
\texttt{~~~dim\_feedforward=128, } \\
\texttt{~~~num\_layers=2)} \\ 
\\
\textbf{Policy:} \\
\texttt{GRU(464,256)} \\
\texttt{ReLU} \\
\texttt{GRU(256,256)} \\
\texttt{Linear(976,4)}
} &

\makecell[l]{
\texttt{Linear(39,1600)} \\
\texttt{ReLU} \\
\texttt{Linear(1600,400)} \\
\texttt{ReLU} \\
\texttt{GRU(400,256)} \\
\texttt{ReLU} \\
\texttt{GRU(256,256)} \\
\texttt{Linear(912,3) $\times$ 3}
} &

\makecell[l]{
\texttt{Linear(39,1600)} \\
\texttt{ReLU} \\
\texttt{Linear(1600,400)} \\
\texttt{ReLU} \\
\texttt{GRU(400,256)} \\
\texttt{ReLU} \\
\texttt{GRU(256,256)} \\
\texttt{Linear(912,24)}
} \\
\bottomrule
\end{tabular}
\end{table}

\label{sec:implementation_details}
All approaches involve supervised training on $\mathcal{D}$, and Ground-Compose-Reinforce and Bespoke Reward Model additionally require an RL phase in the environment. Network architectures for supervised training on $\mathcal{D}$ are reported in Table~\ref{tab:networks1}. PPO network architectures are reported in Table~\ref{tab:networks2}. All policy networks (whether trained via RL or behaviour cloning) use GRUs to temporal dependencies except for GCR, which uses RM transitions. The policy network outputs a probability distribution over actions (in DrawerWorld, the outputs of the network parameterize a Gaussian policy's mean and standard deviation). For methods relying on potential-based reward shaping, we computed shaping rewards without discounting the next potential, i.e. we issued the shaping reward as $\lambda(v' - v)$ rather than $\lambda(\gamma v' - v)$. Though this loses some theoretical convergence properties, we found it to significantly outperform the standard shaping reward in all cases.

\textbf{Ground-Compose-Reinforce.} GCR consists of the following neural networks: a labelling function network that outputs a single binary classification logit for each proposition in $\mathcal{AP}$, a PVF network that outputs an optimal value prediction for each literal in $\mathcal{AP}$, and a policy of the form $\pi(a_t | s_t, u_t)$ that conditions on the current RM state. The labelling function is trained via a binary cross entropy loss on $\mathcal{D}$, the PVFs are trained via offline RL on $\mathcal{D}$, and the policy is trained via RL supported by the labelling function and PVFs to provide learning signals. 

In GeoGrid, PVFs were trained using Fitted Q-Iteration \citep{ernst2005tree}. In DrawerWorld, PVFs were trained to directly predict Monte Carlo returns. We also considered a state-of-the-art offline RL method, MCQ \citep{lyu2022mildly}, but it performed worse than Monte Carlo regression. We attribute this to the relatively small size of $\mathcal{D}$ compared to standard offline RL benchmarks.

\textbf{LTL-conditioned Behaviour Cloning.} This baseline models a neural network policy $\pi_\theta(a_t | h_t, \varphi)$, where the history $h_t$ is encoded by a GRU \citep{cho2014learning} and $\varphi$ (a goal represented directly in LTL) is encoded by a Transformer \citep{vaswani2017attention}. Only observations (and not actions) are encoded as part of the history. We trained $\pi_\theta$ by labelling each trajectory $\tau^i$ in $\mathcal{D}$ with an LTL formula $\varphi^i$ based on the sequence of propositional labels $\sigma^i$ in $\mathcal{D}$ and then minimized the behaviour cloning loss $\mathbb{E}_{\mathcal{D}}[-\log{\pi_\theta(a_t | h_t, \varphi)}]{}$. Finally, we evaluated the model on the downstream tasks by conditioning on the LTL formulas shown in Table~\ref{tab:ltl_formulas}. 

To the best of our knowledge, there are no existing approaches that generate LTL descriptions based on \emph{a single trajectory}. We instead used a custom approach based on common specification templates to generate diverse LTL descriptions. For each trajectory $\tau$ in $\mathcal{D}$, we randomly generated a single formula that is satisfied by $\tau$ for each of the following specification templates found in Table 2 of \citet{8859226}: \emph{visit}, \emph{sequenced visit}, \emph{ordered visit}, \emph{patrolling} (for this purpose, we consider an event to occur infinitely often if it occurs at least five times within the same trajectory in GeoGrid or 200 times within the same trajectory in DrawerWorld) and \emph{global avoidance}. These templates were chosen since they correspond to LTL properties that are relatively simple to automatically mine from a given trajectory. We then labelled each trajectory $\tau$ with a randomly chosen LTL formula from among this set.

\textbf{Bespoke Reward Model.} This baseline is a single neural network that directly predicts the reward, termination, and optimal value function for each of the downstream tasks. The neural network consists of an observation encoder, followed by two GRU layers (to encode the history of observations), followed by three linear output heads to predict rewards, optimal values, and terminations, respectively, for all downstream tasks simultaneously for that domain. To generate target rewards and terminations for a trajectory $\tau$ in $\mathcal{D}$, we evaluated the RM of each downstream task based on the sequence of propositional labels $\sigma^i$. The optimal value estimates were trained using offline RL in a similar manner as the PVFs. Finally, a policy was obtained using RL on the rewards and terminations, while the optimal values were used for potential-based reward shaping, similar to the shaped version of GCR.

\textbf{Bespoke Behaviour Cloning.} This baseline is similar to LTL-BC, except it does not condition on an LTL task---instead, it simultaneously outputs actions for each of the possible downstream tasks. To evaluate the policy on a specific downstream task, only the output for that task is considered. We trained the policy via behaviour cloning on any trajectory that achieves positive return on a particular downstream task due to the limited number of successful demonstrations. 

\begin{table}[t]
\small
\centering
\caption{Network Architectures for PPO.}
\label{tab:networks2}
\setlength{\tabcolsep}{2pt}
\begin{tabular}{@{}cc@{\hskip 1cm}cc@{}}

\cmidrule(r){1-2}\cmidrule{3-4}
\multicolumn{2}{c}{\textbf{GeoGrid}} & \multicolumn{2}{c}{\textbf{DrawerWorld}} \\
\cmidrule(r){1-2}\cmidrule{3-4}

\textbf{GCR} & \textbf{Bespoke Reward Model} &
\textbf{GCR} & \textbf{Bespoke Reward Model} \\

\cmidrule(r){1-2}\cmidrule{3-4}

\makecell[l]{
\textbf{Encoder:} \\
\texttt{Conv2d(6,16,3,1,1)} \\
\texttt{ReLU} \\
\texttt{Conv2d(16,32,3,1,1)} \\
\texttt{ReLU} \\
\texttt{Flatten} \\
\\
\textbf{Actor Head:} \\
\texttt{Linear(2048+$|\mathcal{U}|$,128)} \\
\texttt{ReLU} \\
\texttt{Linear(128,64)} \\
\texttt{ReLU} \\
\texttt{Linear(64,4)} \\
\\
\textbf{Critic Head:} \\
\texttt{Linear(2048+$|\mathcal{U}|$,128)} \\
\texttt{ReLU} \\
\texttt{Linear(128,64)} \\
\texttt{ReLU} \\
\texttt{Linear(64,1)}
} &

\makecell[l]{
\textbf{Encoder:} \\
\texttt{Conv2d(6,16,3,1,1)} \\
\texttt{ReLU} \\
\texttt{Conv2d(16,32,3,1,1)} \\
\texttt{ReLU} \\
\texttt{Flatten} \\
\texttt{GRU(2048,128)} \\
\\
\textbf{Actor Head:} \\
\texttt{Linear(128,128)} \\
\texttt{ReLU} \\
\texttt{Linear(128,64)} \\
\texttt{ReLU} \\
\texttt{Linear(64,4)} \\
\\
\textbf{Critic Head:} \\
\texttt{Linear(128,128)} \\
\texttt{ReLU} \\
\texttt{Linear(128,64)} \\
\texttt{ReLU} \\
\texttt{Linear(64,1)}
} &

\makecell[l]{
\textbf{Actor:} \\
\texttt{Linear(78+$|\mathcal{U}|$,512)} \\
\texttt{ReLU} \\
\texttt{Linear(512,512)} \\
\texttt{ReLU} \\
\texttt{Linear(512,512)} \\
\texttt{ReLU} \\
\texttt{Linear(512,8)} \\
\\
\textbf{Critic:} \\
\texttt{Linear(78+$|\mathcal{U}|$,512)} \\
\texttt{ReLU} \\
\texttt{Linear(512,512)} \\
\texttt{ReLU} \\
\texttt{Linear(512,512)} \\
\texttt{ReLU} \\
\texttt{Linear(512,1)}
} &

\makecell[l]{
\textbf{Encoder:} \\
\texttt{Linear(78,512)} \\
\texttt{ReLU} \\
\texttt{Linear(512,512)} \\
\texttt{GRU(512,512)} \\
\\
\textbf{Actor Head:} \\
\texttt{Linear(512,512)} \\
\texttt{ReLU} \\
\texttt{Linear(512,8)} \\
\\
\textbf{Critic Head:} \\
\texttt{Linear(512,512)} \\
\texttt{ReLU} \\
\texttt{Linear(512,1)}
} \\

\cmidrule(r){1-2}\cmidrule{3-4}

\end{tabular}
\end{table}

\begin{table}[t]
    \small 
    \centering
    \caption{LTL formulas used to evaluate LTL-BC.}
    \label{tab:ltl_formulas}
    \begin{tabular}{l >{\scriptsize}p{0.7\linewidth}}
        \toprule
        \textbf{Task} & \small\textbf{LTL Formula} \\
        \midrule
        \multicolumn{2}{l}{\emph{GeoGrid} (propositions are named r,g,b,c,t instead of $\mathrm{\textcolor{red}{R}}, \mathrm{\textcolor{green}G}, \mathrm{\textcolor{blue}B}, \triangle, \Circle$ to avoid confusion with LTL operators)} \\
        \midrule
        Sequence & $\Diamond\bigl((r \wedge t) \wedge \Diamond((g \wedge t) \wedge \Diamond(b \wedge t))\bigr)$ \\ 
        Loop & $\Box\Diamond\bigl((r \wedge t) \wedge \Diamond((g \wedge t) \wedge \Diamond(b \wedge t))\bigr)$ \\ 
        Logic & $\Diamond(r \wedge t) \wedge
        \Diamond(g \wedge t) \wedge
        \Diamond(b \wedge t) \wedge
        \Diamond(r \wedge c) \wedge
        \Diamond(g \wedge c) \wedge
        \Diamond(b \wedge c) \wedge
        (\neg b \, \mathcal{U} \, r) \wedge
        (\neg g \, \mathcal{U} \, b)
        $ \\
        Safety & $\Diamond\bigl((r \wedge t) \wedge \Diamond((g \wedge t) \wedge \Diamond(b \wedge t))\bigr)
        \wedge \Box \neg t$ \\ 
        \midrule 
        \multicolumn{2}{l}{\emph{DrawerWorld}} \\
        \midrule
        Hold-Red-Box & $\Box\Diamond \mathrm{RedBoxLifted}$ \\
        Pickup-Each-Box & $\Diamond (\mathrm{RedBoxLifted} \wedge \Diamond (\mathrm{BlueBoxLifted} \wedge \Diamond \mathrm{GreenBoxLifted}))$ \\
        Show-Green-Box & $ \neg [\neg (\mathrm{GreenBoxInDrawer1} \wedge \Diamond(\mathrm{Drawer1Open} \, \mathcal{U} \, \mathrm{GreenBoxLifted}))$ \\
        & ~~~$\wedge \neg (\mathrm{GreenBoxInDrawer2} \wedge \Diamond(\mathrm{Drawer2Open} \, \mathcal{U} \, \mathrm{GreenBoxLifted}))$ \\
        & ~~~$\wedge \neg (\neg \mathrm{GreenBoxInDrawer1} \wedge \neg \mathrm{GreenBoxinDrawer2} \wedge \Diamond \mathrm{GreenBoxLifted})] $ \\
        \bottomrule
    \end{tabular}
\end{table}


\subsection{Experimental Setup and Hyperparameter Details}
\label{sec:training_details}

\paragraph{Details: Supervised Training on $\mathcal{D}$.}
All experiments were run on a compute cluster. Supervised training on $\mathcal{D}$ required a single GPU and CPU, minimal memory resources (24GB of RAM or less) and no more than 30 minutes to train any method to 100 epochs. We tuned hyperparameters via a line search over batch size, learning rate, L1 regularization coefficient, and epochs (in that order) using a held-out 10\% of the trajectories in D, and the final hyperparameters are reported in Table~\ref{tab:hyperparams}. Final models were retrained on the full data with the tuned hyperparameters. We note that the batch size hyperparameter should be interpreted differently for methods requiring a GRU. For GCR, it refers to the number of transitions sampled from $\mathcal{D}$. For LTL-BC, Bespoke Reward Model and Bespoke Behaviour Cloning, it refers to the number of full length \emph{trajectories} sampled from $\mathcal{D}$. This is because it is necessary to keep transitions in a trajectory in the correct order to train the GRU.

\begin{table}[t]
\footnotesize
\centering
\caption{Hyperparameters for Supervised Training on $\mathcal{D}$.}
\label{tab:hyperparams}
\begin{tabular}{lcccccc}
\toprule
\textbf{Hyperparameter} 
& \multicolumn{2}{c}{\textbf{GCR}} 
& \textbf{LTL-BC} 
& \textbf{Bespoke Reward Model} 
& \textbf{Bespoke BC} \\
& {\scriptsize Labelling Function} & {\scriptsize PVFs} & & & \\
\midrule
\multicolumn{6}{l}{\textit{GeoGrid}} \\
\midrule
Batch size            & 256     & 1024    & 100  & 100   & 100   \\
Learning rate         & 3e-4    & 3e-4    & 1e-4  & 3e-3  & 3e-4  \\
L1 loss   & 1e-5    & 0       & 0  & 0     & 0     \\
Epochs                & 10      & 100     & 9  & 100   & 11    \\
Discount factor       & n.a.    & 0.97    & n.a.  & 0.97  & n.a.  \\
\midrule
\multicolumn{6}{l}{\textit{DrawerWorld}} \\
\midrule
Batch size            & 256     & 256     & 50  & 50    & 50    \\
Learning rate         & 3e-4    & 3e-4    & 1e-3  & 1e-4  & 1e-4  \\
L1 loss   & 1e-5    & 1e-5    & 0  & 0     & 1e-9  \\
Epochs                & 100     & 100     & 87  & 11    & 4     \\
Discount factor       & n.a.    & 0.9975  & n.a.  & 0.9975 & n.a. \\
\bottomrule
\end{tabular}
\end{table}

\paragraph{Details: RL Training.} 

\begin{table}[h!]
\small
\centering
\caption{RL Training Hyperparameters.}
\label{tab:ppo_hyperparams}
\begin{tabular}{lc}
\toprule
\textbf{Hyperparameter} & \textbf{Value for all methods} \\
\midrule
\multicolumn{2}{l}{\textit{GeoGrid}} \\
\midrule
Number of parallel environments & 16 \\
Frames per update per process & 1000 \\
Learning rate        & 3e-4         \\
Discount factor ($\gamma$)  & 0.97         \\
GAE parameter ($\lambda$)   & 0.95         \\
Clip range                  & 0.2          \\
Entropy coefficient         & 1e-4         \\
Value loss coefficient      & 0.5          \\
Number of epochs per update & 4           \\
Minibatch size              & 4000           \\
High-level Discount Factor ($\gamma_\mathrm{RM}$) & 0.97$^\text{10}$ \\
Shaping potential weighting coefficient ($\lambda$) & 1 \\
\midrule
\multicolumn{2}{l}{\textit{DrawerWorld}} \\
\midrule
Number of parallel environments & 16 \\
Frames per update per process & 4000 \\
Learning rate        & 3e-4         \\
Discount factor ($\gamma$)  & 0.99         \\
GAE parameter ($\lambda$)   & 0.99         \\
Clip range                  & 0.2          \\
Entropy coefficient         & 0.01 in Pickup-Each-Box, otherwise 0.03         \\
Value loss coefficient      & 0.5          \\
Number of epochs per update & 10           \\
Minibatch size              & 8000           \\
High-level Discount Factor ($\gamma_\mathrm{RM}$) & 0.9975$^\text{400}$ \\
Shaping potential weighting coefficient ($\lambda$) & 0.1 in Hold-Red-Box, otherwise 1 \\
\bottomrule
\end{tabular}
\end{table}

All experiments were run on a compute cluster. Each RL run used a single GPU, 16 CPUs, and 48GB of RAM. For GCR, runs took up to 6 hours on GeoGrid (to train to 15M frames) and 16 hours on DrawerWorld (to train to 20M frames). RL training with the Bespoke Reward Model took longer due to GRUs---up to 12 hours on GeoGrid (to train to 15M frames) and 18 hours on DrawerWorld (to train to 20M frames). For RL training, we used the implementation of PPO at \url{https://github.com/lcswillems/torch-ac} with the hyperparameters in Table~\ref{tab:ppo_hyperparams}. The total number of environment steps each method was trained on was 2.5M for Sequence, 4M for Loop, 10M for Safety, 20M for Show-Green-Box, and 15M for all others. 



\subsection{Learning Curves}
\label{sec:learning_curves}

\begin{figure}[t]
    \centering
    \includegraphics[width=0.42\textwidth]{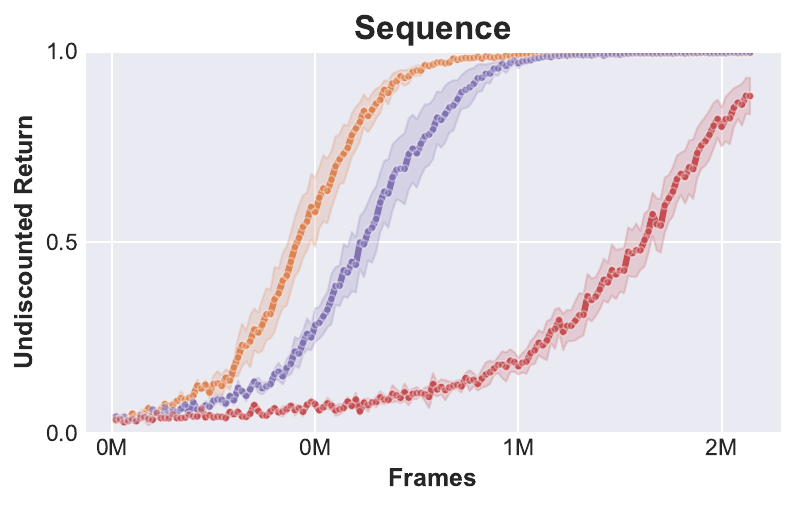}
    \includegraphics[width=0.43\textwidth]{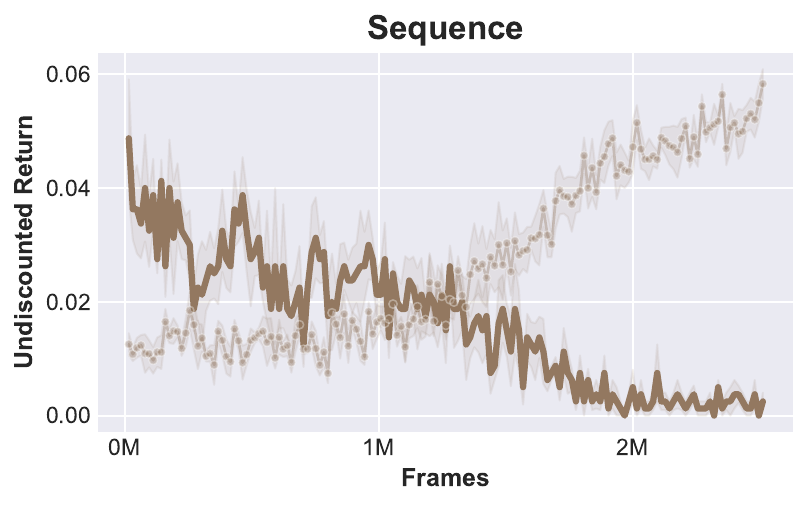}
    \includegraphics[width=0.42\textwidth]{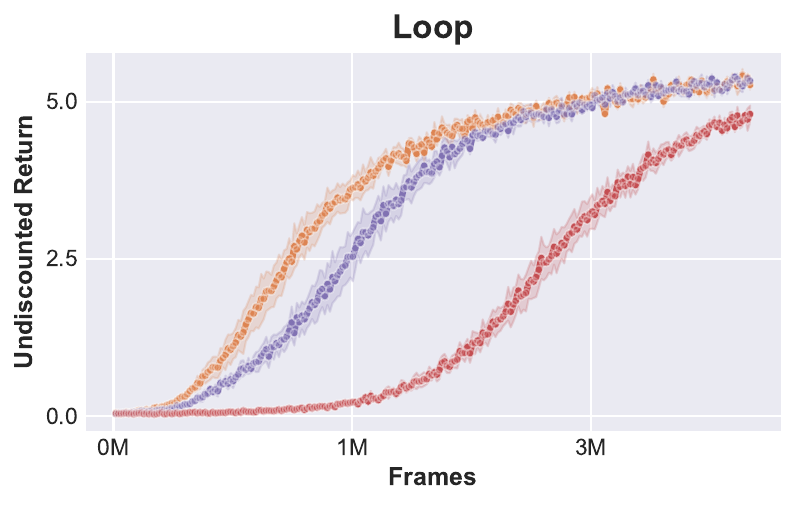}
    \includegraphics[width=0.44\textwidth]{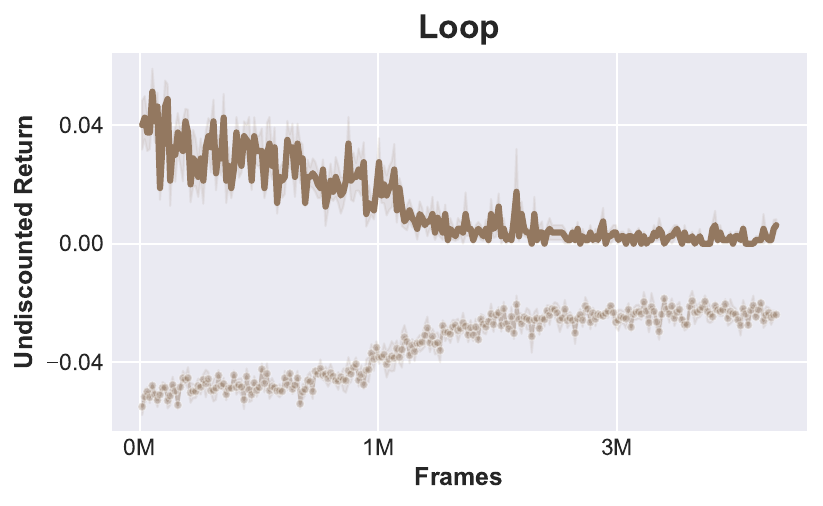}
    \includegraphics[width=0.42\textwidth]{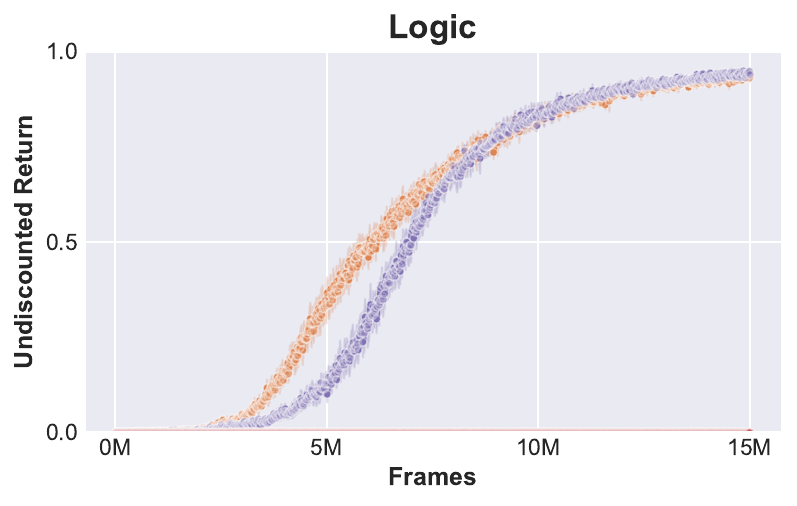}
    \includegraphics[width=0.45\textwidth]{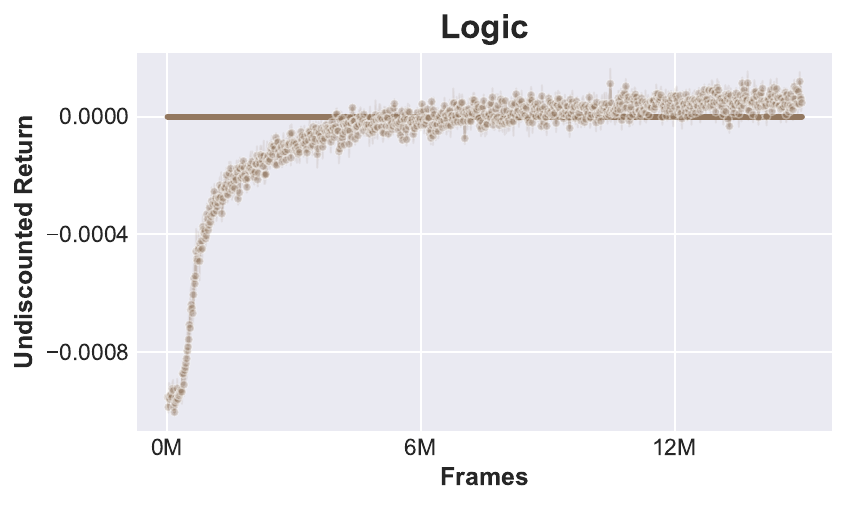}
    \includegraphics[width=0.42\textwidth]{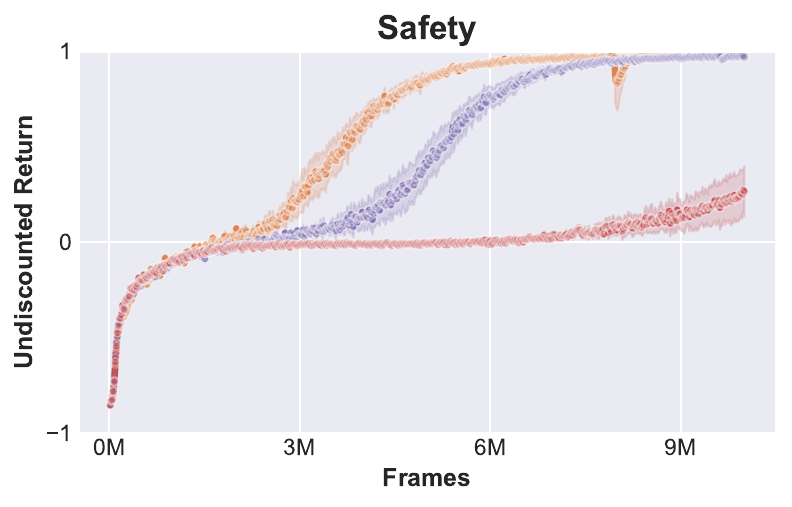}
    \includegraphics[width=0.44\textwidth]{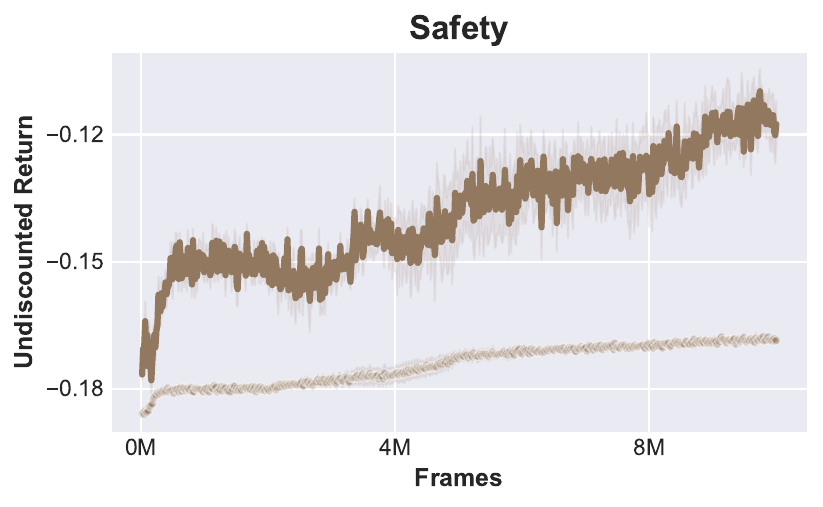}

    \includegraphics[width=0.75\textwidth]{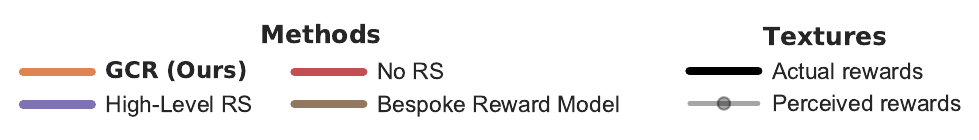}
    \caption{RL learning curves for GeoGrid, showing returns under the agent's own reward model (``perceived rewards'') and under the ground-truth $\mathcal{L}^*$ (``actual rewards''). Perceived rewards are reported without shaped rewards, and shaded regions show standard error. Approaches based on Ground-Compose-Reinforce (including No RS and High-Level RS) generate rewards that are closely aligned with the ground truth and lead to an effective final policy, while Bespoke Reward Model produces near-zero rewards in all cases and makes little progress.
    }
    \label{fig:learning_curves1}
\end{figure}

\begin{figure}[t]
    \centering
    \includegraphics[width=0.42\textwidth]{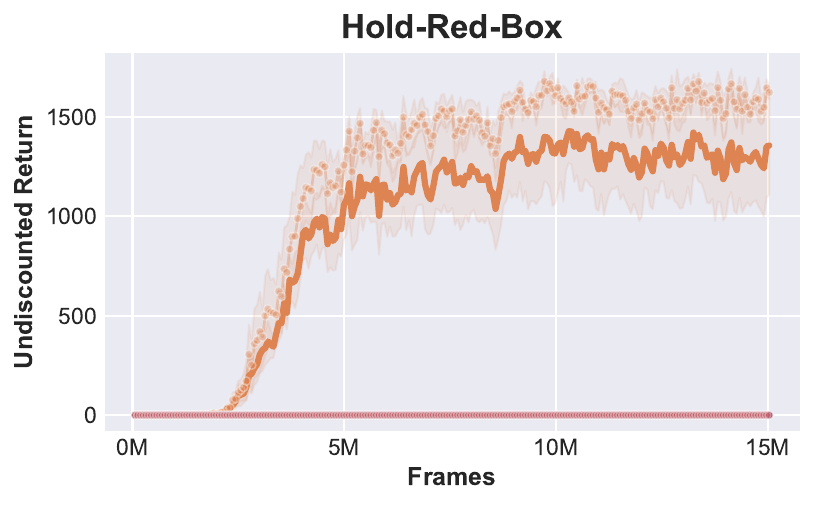}
    \includegraphics[width=0.42\textwidth]{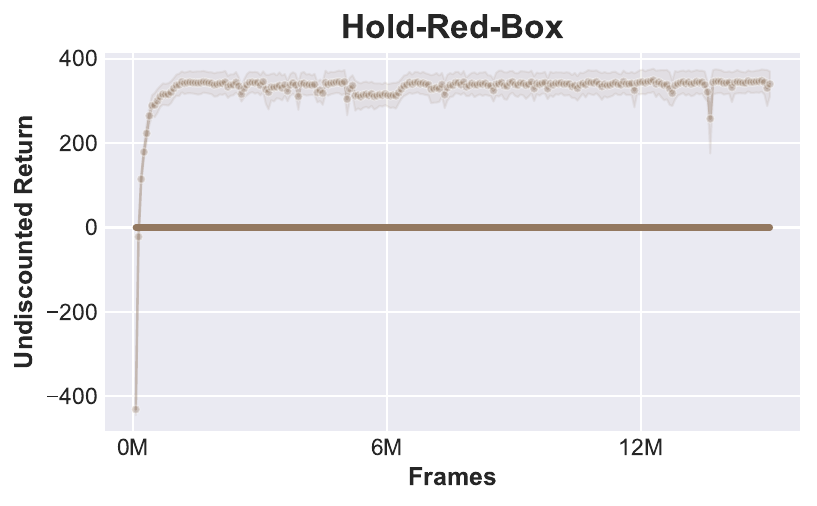}
    \includegraphics[width=0.42\textwidth]{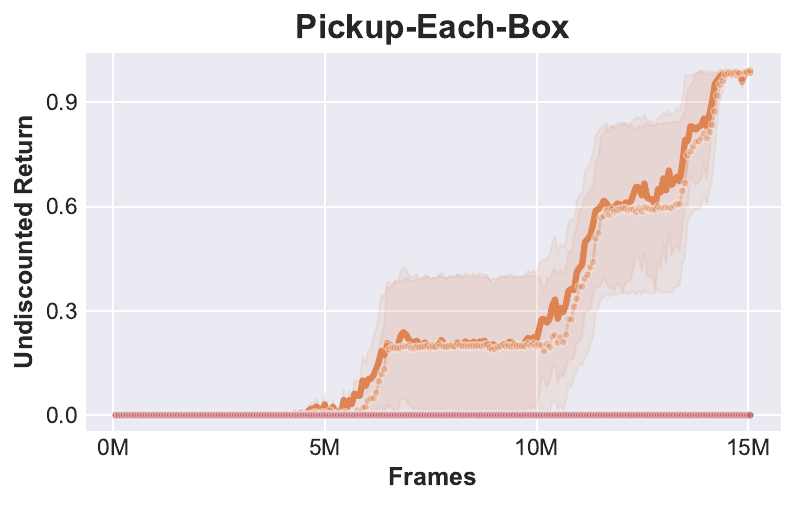}
    \includegraphics[width=0.42\textwidth]{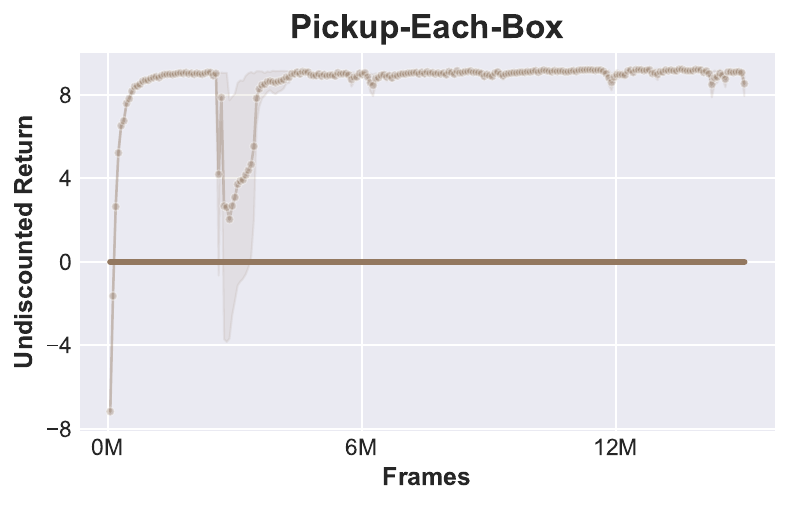}
    \includegraphics[width=0.42\textwidth]{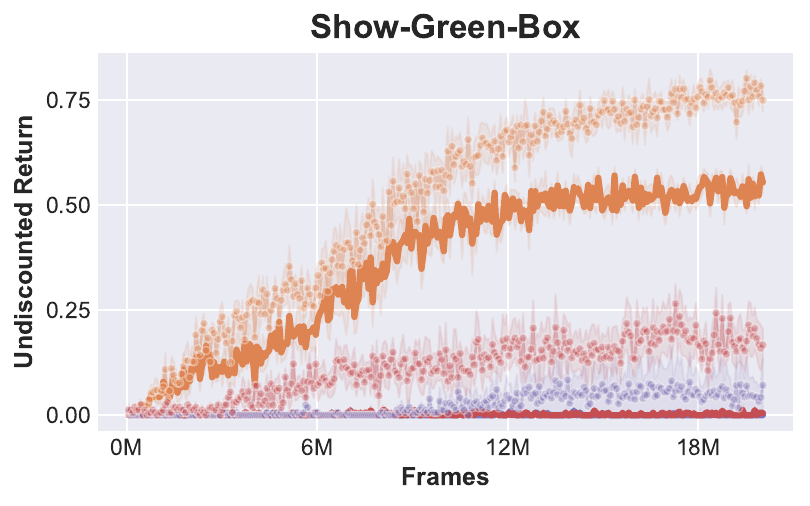}
    \includegraphics[width=0.42\textwidth]{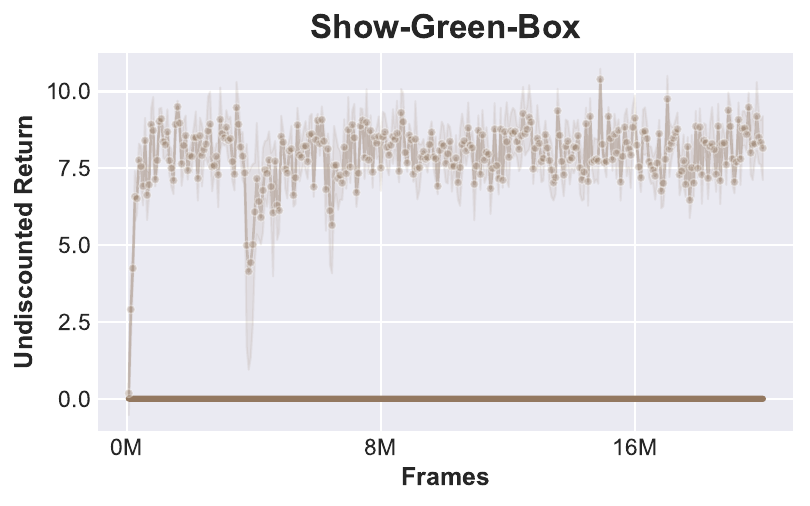}
    
    \includegraphics[width=0.75\textwidth]{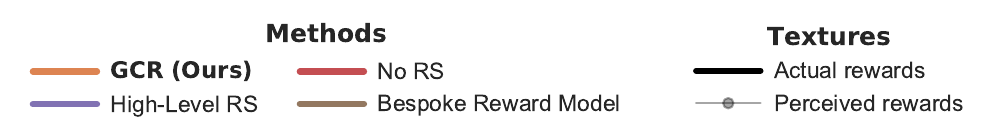}
    
    \caption{RL learning curves for DrawerWorld, showing returns under the agent's own reward model (``perceived rewards'') and under the ground-truth $\mathcal{L}^*$ (``actual rewards''). Perceived rewards are reported without shaped rewards, and shaded regions show standard error. When evaluated under ground-truth rewards, Ground-Compose-Reinforce with our reward shaping strategy learns strong policies in all cases, while alternative approaches make no progress. Notably, Bespoke Reward Model results in a final policy with high \emph{perceived rewards}, but poor actual performance. }
    \label{fig:learning_curves2}
\end{figure}

We report RL learning curves for each method and task in Figures~\ref{fig:learning_curves1} and \ref{fig:learning_curves2}. 
GCR and its variants have an internal reward model that is better aligned with the ground truth compared to Bespoke Reward Model. In GeoGrid, GCR's reward model is near perfect, and in all cases, optimizing its internal rewards improves ground truth performance as well. Bespoke Reward Model almost always predicts near-zero rewards on GeoGrid tasks (except Safety), likely since there are few examples of positive demonstrations in $\mathcal{D}$. On DrawerWorld, Bespoke Reward-Model is highly misaligned, predicting large rewards but garnering near-zero return based on the ground truth. 

In terms of sample efficiency, we observe that GCR with our reward shaping strategy outperforms all baselines. The difference is marginal in GeoGrid, where exploration is less of an issue, but in DrawerWorld, all other reward shaping approaches fail.

\section{Autoformalizing Natural Language to Reward Machines}
\label{sec:autoformalization}

While several works have been dedicated to the autoformalization of natural language instructions into formal languages such as LTL \citep{brunello2019synthesis, liu2022lang2ltl, fuggitti2023nl2ltl, chen2023nl2tl}, we show that a modern LLM can sometimes perform this task zero-shot for RMs, without having specifically been trained on it (to the best of our knowledge). This allows us to directly task the RL agent in our framework through natural language, then autoformalize the task description into an RM.

We tested OpenAI's ChatGPT-4o and o3 models as the autoformalizer and prompted it with a description of the autoformalization task (including the output format), a text description of the environment, a list of propositions and associated text descriptions, and a text description of the desired reward function (Listing~\ref{lst:sample}). For each of the four GeoGrid tasks in Table~\ref{tab:tasks}, we ran each autoformalizer five times for consistency. We manually evaluated each outputted RM based on whether it yielded a reward function that exactly matched the textual description, with the success rate reported in Table~\ref{tab:autoformalization_results}. 

ChatGPT-4o correctly produced RMs for all tasks, except for Logic, which requires a complex RM (our solution involved 10 states and 18 transitions). However, we note that ChatGPT-4o was nearly correct for all five trials for Logic---each of its outputted RMs deviated by a single transition that changed the behaviour of the resultant reward function. o3 outputted correct RMs on all tasks. Notably, the outputted RMs were identical in structure to the intended RMs we manually constructed in all cases (with the only differences being in the naming of RM states and the representation of equivalent logical formulas).  

\begin{minipage}[t]{\textwidth}
\begin{lstlisting}[caption=Reward Machine Autoformalization Prompt,label=lst:sample]
You are given a list of propositional symbols and their descriptions, an environment description, and a description of a desired reward function in English. Your job is to construct a Reward Machine representing this reward function.

Reward Machine states should be numbered 0, 1, 2, 3, ..., with 1 always being the initial state, and 0 always being the terminal state. Transitions should be represented as a tuple (i, j, \varphi, r), where i and j are the start and end Reward Machine states of the transition, respectively, \varphi is a logical formula over the set of propositional symbols (use "!" to represent "not", "&" to represent "and", and "|" to represent "or", and write the formula in disjunctive normal form), and r is the reward for the transition. Your output should be the transitions in the Reward Machine, one per line, in the tuple form shown above, e.g. (0, 1, !X&Y, 0.1), with no other punctuation. For brevity, do not list self-loop transitions that provide 0 reward in the output.

Environment Description: The environment is a gridworld, where some squares have objects. Each object has a single colour (red, green, or blue) and a single shape (circle, or triangle). 

Propositions:
- red: The agent's current cell has a red object.
- blue: The agent's current cell has a blue object.
- green: The agent's current cell has a green object.
- triangle: The agent's current cell has a triangle.
- circle: The agent's current cell has a circle.

Task: <TASK DESCRIPTION>
\end{lstlisting}
\end{minipage}

\begin{table}[t]
\small
\centering
\caption{Success rate for ChatGPT-4o and o3 when autoformalizing Reward Machines from a natural language description of the desired reward function.}
\label{tab:autoformalization_results}
\renewcommand{\arraystretch}{1.2}
\begin{tabular}{lp{6cm}cc}
\hline
\textbf{Task} & \textbf{Description} & \textbf{GPT-4o Success Rate} & \textbf{o3 Success Rate}  \\
\hline
Sequence & Give a reward of 1 and terminate the episode when a red triangle, a green triangle, and a blue triangle have been reached, in that order. Only give the reward of 1 when the final step has been completed. Give 0 reward and never terminate the episode otherwise. & 100\% & 100\% \\
\hline
Loop & Give a reward of 1 when a red triangle, a green triangle, and a blue triangle have been reached, in that order. Only give the reward of 1 when the final step has been completed. After completing this sequence, the agent may repeat all steps of the sequence to receive the reward again, as many times as it wishes. The episode never terminates. & 100\% & 100\% \\
\hline
Logic & Give a reward of 1 and terminate the episode as soon as a red triangle, red circle, green triangle, green circle, blue triangle, and blue circle have all been reached at some point. However, blue objects should not be visited until both red objects are visited, and green objects should not be visited until both blue objects are visited. Circles and triangles of the same colour can be reached in either order. If any objects are reached out of order, immediately terminate the episode with a reward of 0.
 & 0\% & 100\% \\
\hline
Safety & Give a reward of 1 when a red object, a green object, and a blue object have been reached, in that order. Only give the reward of 1 when the final step has been completed. However, always avoid squares with triangles---if this is violated, immediately terminate the episode with a reward of -1. & 100\% & 100\% \\ 
\hline
\end{tabular}
\end{table}

\section{Societal Impact}
\label{sec:societalimpact}

\textbf{Data‑efficient learning lowers environmental cost.}
Ground‑Compose‑Reinforce (GCR) achieves strong generalization from a relatively small, task‑agnostic trajectory dataset. Because it avoids the need for internet‑scale demonstrations, the total compute and data collection burden is substantially reduced, which in turn diminishes the carbon footprint of training and retraining large decision‑making systems.

\textbf{Transparent, verifiable task specifications.}
By exposing an explicit formal specification layer (Reward Machines) between the human and the agent, GCR allows auditors to read, simulate, and formally verify the reward logic before deployment. This contrasts with opaque end‑to‑end reward models and can help regulators trace undesirable behaviour back to a concrete symbolic condition rather than a latent neural representation, supporting safer and more accountable RL pipelines.

\textbf{Broader access to capable agents.}
Because symbols are grounded once and then recomposed to create an unbounded task space, domain experts without ML backgrounds can author complex tasks simply by writing RMs, 
potentially democratizing advanced robotics and simulation tools in education, manufacturing, and assistive settings. The same mechanism lets small‑lab researchers prototype complex multi‑stage tasks without the costs associated with collecting new labelled rewards.

\textbf{Reward hacking and specification gaps.}
If the learned interpretation of propositions in the environment is erroneous, the resultant behaviour may no longer match human intent. In the paper, we caution that such mis-grounding can lead to harmful behaviours despite the use of a precise formal specification.


\textbf{Labour displacement.}
Easier programming of general‑purpose robotic skills may substitute for manual labour in logistics or assembly lines, contributing to job displacement without adequate social safety nets.

%

\end{document}